\documentclass[runningheads]{llncs}

\usepackage{microtype}
\usepackage{subfigure}
\usepackage{booktabs} 
\usepackage{mathtools}  
\usepackage{amsfonts}  
\usepackage{amsmath}
\newcommand{\xv}{\mathbf{x}}
\newcommand{\mli}[1]{\mathit{#1}}
\usepackage{verbatim}
\usepackage{fancyvrb}  
\usepackage{hyperref}
\usepackage[table]{xcolor}
\usepackage{float}
\usepackage{todonotes}
\usepackage{wrapfig}
\makeatletter 
\newcommand\nobreakpar{\par\nobreak\@afterheading} 
\makeatother

\usepackage{graphicx}
\begin{document}
\title{Multi-Objective Counterfactual Explanations\thanks{This work has been partially supported by the German Federal Ministry of Education and Research (BMBF) under Grant No. 01IS18036A and by the Bavarian State Ministry of Science and the Arts in the framework of the Centre Digitisation.Bavaria (ZD.B). The authors of this work take full responsibility for its content.}}

\author{Susanne Dandl\orcidID{0000-0003-4324-4163} \and
Christoph Molnar\orcidID{0000-0003-2331-868X} \and
Martin Binder \and
Bernd Bischl\orcidID{0000-0001-6002-6980}}
\authorrunning{S. Dandl et al.}
%
\institute{Department of Statistics, LMU Munich, Ludwigstr. 33, 80539 Munich, Germany \email{susanne.dandl@stat.uni-muenchen.de}}
\maketitle              
\begin{abstract}
Counterfactual explanations are one of the most popular methods to make predictions of black box machine learning models interpretable by providing explanations in the form of `what-if scenarios'. 
Most current approaches optimize a collapsed, weighted sum of multiple objectives, which are naturally difficult to balance a-priori.
We propose the Multi-Objective Counterfactuals (MOC) method, which translates the counterfactual search into a multi-objective optimization problem. Our approach not only returns a diverse set of counterfactuals with different trade-offs between the proposed objectives, but also maintains diversity in feature space. This enables a more detailed post-hoc analysis to facilitate better understanding and also more options for actionable user responses to change the predicted outcome. Our approach is also model-agnostic and works for numerical and categorical input features. We show the usefulness of MOC in concrete cases and compare our approach with state-of-the-art methods for counterfactual explanations. 
\end{abstract} 

\section{Introduction}
\label{introduction}

Interpretable machine learning methods have become very important in recent years to explain the behavior of black box machine learning (ML) models. 
A useful method for explaining \textit{single} predictions of a model are counterfactual explanations. 
ML credit risk prediction is a common motivation for counterfactuals.
For people whose credit applications have been rejected, it is valuable to know why they have not been accepted, either to understand the decision making process or to assess their actionable options to change the outcome.
Counterfactuals provide these explanations in the form of ``if these features had different values, your credit application would have been accepted". 
For such explanations to be plausible, they should only suggest small changes in a few features. 
Therefore, counterfactuals can be defined as close neighbors of an actual data point, but their predictions have to be sufficiently close to a (usually quite different) desired outcome. 
Counterfactuals explain why a certain outcome was not reached, can offer potential reasons to object against an unfair outcome and give guidance on how the desired prediction could be reached in the future \cite{wachter17}. 
Note that counterfactuals are also valuable for predictive modelers on a more technical level to investigate the pointwise robustness and the pointwise bias of their model.

\section{Related Work}
\label{relatedwork}
Counterfactuals are closely related to adversarial perturbations. These have the aim to deceive ML models instead of making the models interpretable \cite{su17}. 
Attribution methods such as Local Interpretable Model-agnostic Explanations (LIME) \cite{ribeiro16} and Shapley Values \cite{lundberg17} explain a prediction by determining how much each feature contributed to it.
Counterfactual explanations differ from feature attributions since they generate data points with a different, desired prediction instead of attributing a prediction to the features. 

Counterfactual methods can be model-agnostic or model-specific. The latter usually exploit the internal structure of the underlying ML model, such as the trained weights of a neural network, while the former are based on general principles which work for arbitrary ML models - often by only assuming access to the prediction function of an already fitted model.
Several model-agnostic counterfactual methods have been proposed \cite{dhurandhar19,grath18,karimi19,laugel17,poyiadzi19,sharma19,white19}.
Apart from Grath et al. \cite{grath18}, these approaches are limited to classification.
Unlike the other methods, the method of Poyiadzi et al. \cite{poyiadzi19} can obtain plausible counterfactuals by constructing feasible paths between data points with opposite predictions. 

A model-specific approach was proposed by Wachter et al. \cite{wachter17}, who also introduced and formalized the concept of counterfactuals in predictive modeling. 
Like many model-specific methods \cite{joshi19,looveren19,mothilal19,russell19,ustun19} their approach is limited to differentiable models.
The approach of Tolomei et al. \cite{tolomei17} generates explanations for tree-based ensemble binary classifiers. As with \cite{wachter17} and \cite{looveren19}, it only returns a single counterfactual per run. 

\section{Contributions}
\label{contribution}
In this paper, we introduce Multi-Objective Counterfactuals (MOC), which to the best of our knowledge is the first method to formalize the counterfactual search as a multi-objective optimization problem. 
We argue that the mathematical problem behind the search for counterfactuals should be naturally addressed as multi-objective. 
Most of the above methods optimize a collapsed, weighted sum of multiple objectives to find counterfactuals, which are naturally difficult to balance a-priori. 
They carry the risk of arbitrarily reducing the solution set to a single candidate without the option to discuss inherent trade-offs -- which should be especially relevant for model interpretation that is by design very hard to precisely capture in a (single) mathematical formulation.

Compared to Wachter et al. \cite{wachter17}, we use a distance metric for mixed feature spaces and two additional objectives: one that measures the number of feature changes to obtain sparse and therefore more interpretable counterfactuals, and one that measures the closeness to the nearest observed data points for more plausible counterfactuals. 
MOC returns a Pareto set of counterfactuals that represents different trade-offs between our proposed objectives, and which are constructed to be diverse in feature space.
This seems preferable because changes to different features can lead to a desired counterfactual prediction\footnote{Rashomon effect \cite{breiman01}} and it is more likely that some counterfactuals meet the (hidden) preferences of a user. 
A single counterfactual might even suggest a strategy that is interpretable but not actionable (e.g., `reduce your number of pregnancies') or counterproductive in more general contexts (e.g., `increase your age to reduce the risk of diabetes').
In addition, if multiple otherwise quite different counterfactuals suggest changes to the same feature, the user may have more confidence that the feature is an important lever to achieve the desired outcome.
We refer the reader to Appendix~\ref{ap:examples} for two concrete examples illustrating the above.

Compared to other counterfactual methods, MOC is model-agnostic and handles classification, regression and mixed feature spaces, which furthermore increases its practical usefulness in general applications. Together with \cite{karimi19}, our paper also includes one of the first benchmark studies that compares multiple counterfactual methods on multiple, heterogeneous datasets.

\section{Methodology}
\label{methodology}
\cite{wachter17} loosely define counterfactuals as:
\begin{quote}
\begin{footnotesize}
``You were denied a loan because your annual income was £30,000. If
your income had been £45,000, you would have been offered a loan.”
Here the statement of decision is followed by a counterfactual, or
statement of how the world would have to be different for a desirable
outcome to occur. Multiple counterfactuals are possible, as multiple
desirable outcomes can exist, and there may be several ways to achieve
any of these outcomes."
\end{footnotesize}
\end{quote}
We now formalize this statement by stating four objectives, which a counterfactual should adhere to. In the subsequent section we provide detailed definitions of these objectives and tie them together as a multi-objective optimization problem in order to generate a diverse set of different trade-off solutions.

\subsection{Multi-Objective Counterfactuals}

\begin{definition}[Counterfactual Explanation]
	\label{def:counterfactual}
	Let $\hat{f}:\mathcal{X} \rightarrow \mathbb{R}$ be a prediction function, $\mathcal{X}$ the feature space and $Y' \subset \mathbb{R}$ a set of desired outcomes. The latter can either be a single value or an interval of values. 
	We define a counterfactual explanation $\xv'$ for an observation $\xv^*$ as a data point fulfilling the following: (1) its prediction $f(\xv')$ is close to the desired outcome set $Y'$, (2) it is close to $\xv^*$ in the $\mathcal{X}$ space, (3) it differs from $\xv^*$ only in a few features, and (4) it is a \emph{plausible} data point according to the probability distribution $\mathbb{P}_{\mathcal{X}}$.
For classification models, we assume that $\hat{f}$ returns the probability for a user-selected class and  $Y'$ has to be the desired probability (range).
\end{definition}
This can be translated into a multi-objective minimization task:
\begin{equation}
\label{eq:moo}
\min_\xv \mathbf{o}(\xv) := \min_\xv \big(o_1(\hat{f}(\xv), Y'),\, o_2(\xv, \xv^*), o_3(\xv, \xv^*), o_4(\xv, \mathbf{X}^{obs})\big), 
\end{equation}
with $\mathbf{o}:\mathcal{X} \rightarrow \mathbb{R}^4$ and $\mathbf{X}^{obs}$ as the observed (i.e. training) data. The first component $o_1$ quantifies the distance between $\hat{f}(\xv)$ and $Y'$. We define it as:\footnote{We chose the $L_1$ norm over the $L_2$ norm for a natural interpretation. Its non-differentiability is negligible for evolutionary optimization.}
\begin{equation*}o_1(\hat{f}(\xv), Y') = 
\begin{cases}
0 & \textnormal{if $\hat{f}(\xv) \in Y'$} \\
\inf \limits_{y' \in Y'}|\hat{f}(\xv) - y'| & \textnormal{else}
\end{cases}.
\end{equation*}
The second component $o_2$ quantifies the distance between $\xv^*$ and $\xv$ 
using the Gower distance to account for mixed features \cite{gower71}:
\begin{equation*}
o_2(\xv, \xv^*) = \frac{1}{p}\sum_{j = 1}^{p} \delta_G(x_j, x^*_j)\in [0, 1]
\end{equation*}
with $p$ being the number of features.
The value of $\delta_G$ depends on the feature type:
\begin{equation*}
\delta_G(x_j, x^*_j) = 
\begin{cases}
\frac{1}{\widehat{R}_j}|x_j- x^*_j| & \text{if $x_j$ is numerical} \\
\mathbb{I}_{x_j \neq x_j^*} & \text{if $x_j$ is categorical}
\end{cases}
\end{equation*}
with $\widehat{R}_j$ as the value range of feature $j$, extracted from the observed dataset. 

Since the Gower distance does not take into account how many features have been changed, we introduce objective $o_3$, which counts the number of changed features using the $L_0$ norm:
\begin{equation*}
o_3(\xv, \xv^*) = ||\xv-\xv^*||_0 = \sum_{j = 1}^{p}\mathbb{I}_{x_j\neq x^*_j}.
\end{equation*}
The fourth objective $o_4$ measures the weighted average Gower distance between $\xv$ and the $k$ nearest observed data points $\xv^{[1]}, ..., \xv^{[k]} \in \textbf{X}^{obs}$ as an empirical approximation of how likely $\xv$ originates from the distribution of $\mathcal{X}$:

\begin{equation*}
o_4(\xv, \textbf{X}^{obs}) = \sum_{i = 1}^k w^{[i]} \frac{1}{p} \sum_{j = 1}^{p}  \delta_G(x_j, x^{[i]}_j) \in [0, 1] \text{ where }  \sum_{i = 1}^k w^{[i]} = 1. 
\end{equation*}
Throughout this paper, we set $k$ to 1.
Further procedures to increase the plausibility of the counterfactuals are integrated into the optimization algorithm and are described in Section~\ref{modification}.

Balancing the four objectives is difficult since the objectives contradict each other.
For example, minimizing the distance between counterfactual outcome and desired outcome $Y'$ ($o_1$) becomes more difficult when we require counterfactual feature values close to $\xv^*$ ($o_2$ and $o_3$) and to the observed data ($o_4$). 

\subsection{Counterfactual Search}
\label{nsga}
Our proposed method MOC uses the \emph{Nondominated Sorting Genetic Algorithm II} (NSGA-II) \cite{deb02} with modifications specific to the problem considered. First, unlike the original NSGA-II, it uses \emph{mixed integer evolutionary strategies} (MIES) \cite{li13} to work with the mixed discrete and continuous search space. 
Furthermore, a different crowding distance sorting algorithm is used, and we propose some optional adjustments tailored to the counterfactual search in the upcoming section. 

For MOC, each candidate is described by its feature vector (the `genes') and the objective values of the candidates are evaluated by Eq.~(\ref{eq:moo}). 
Features of candidates are recombined and mutated with predefined probabilities -- some of the control parameters of MOC. 
Numerical features are recombined by the simulated binary crossover recombinator \cite{deb95}, all other feature types by the uniform crossover recombinator \cite{syswerda89}.
Based on \cite{li13}, numerical features are mutated by the scaled Gaussian mutator.
Categorical features are altered by uniformly sampling from their admissible levels, while binary and logical features are simply flipped.
After recombination and mutation, some feature values are randomly set to the values of $\xv^*$ with a given (low) probability -- another control parameter -- to prevent all features from deviating from $\xv^*$.

Contrary to NSGA-II, the crowding distance is computed not only in the objective space $\mathbb{R}^4$ ($L_1$ norm) but also in the feature space $\mathcal{X}$ (Gower distance), and the distances are summed up with equal weighting. 
As a result, candidates are more likely kept if they differ greatly from another candidate in their feature values although they are similar in the objective values.
Diversity in $\mathcal{X}$ is desired because the chances of obtaining counterfactuals that meet the (hidden) preferences of users are higher.
This approach is based on Avila et al. \cite{avila06}. 

MOC stops if either a predefined number of generations is reached (default) or the performance no longer improves for a given number of successive generations. 

\subsection{Further Modifications}
\label{modification}

\subsubsection{Initialization} Naively, we could initialize a population by uniformly sampling some feature values from their full range of possible values, 
while randomly setting other features to the values of $\xv^*$ to induce sparsity.
However, if a feature has a large influence on the prediction, it should be more likely that the counterfactual values differ from $\xv^*$. 
The importance of a feature for an entire dataset can be measured as the standard deviation of the partial dependence plot \cite{greenwell18}. 
Analogously, we propose to measure the feature importance for a single prediction with the standard deviation of the Individual Conditional Expectation (ICE) curve of $\xv^*$. 
ICE curves show for one observation and for one feature how the prediction changes when the feature is changed, while other features are fixed to the values of the considered observation \cite{goldstein15}. The greater the standard deviation of the ICE curve, the higher we set the probability that the feature value is initialized with a different value than the one of $\xv^*$. 
Therefore, the standard deviation $ \sigma^{ICE}_j$ of each feature $x_j$ is transformed into probabilities within $[p_{min}, p_{max}] \cdot 100\%$:
\begin{equation*}
P(\textit{value differs}) = 
\frac{(\sigma^{\mli{ICE}}_j - \mli{min}(\sigma^{\mli{ICE}}))\cdot (p_{max} - p_{min})}{ \mli{max}(\sigma^{\mli{ICE}}) -  \mli{min}(\sigma^{\mli{ICE}})}  + p_{min} 
\end{equation*}
with $\boldsymbol{\sigma}^{ICE} := (\sigma^{ICE}_1, ..., \sigma^{ICE}_p)$. 
$p_{min}$ and $p_{max}$ are control parameters with default values 0.01 and 0.99.

\subsubsection{Actionability} To get more actionable counterfactuals, extreme values of numerical features outside a predefined range are capped to the upper or lower bound after recombination and mutation. 
The ranges can either be derived from the minimum and maximum values of the features in the observed dataset or users can define these ranges. 
In addition, users can identify non-actionable features such as the country of birth or gender. The values of these features are permanently set to the values of $\xv^*$ for all candidates within MOC. 

\subsubsection{Penalization} Furthermore, candidates whose predictions are further away from the target than a predefined distance $\epsilon \in \mathbb{R}$ can be penalized. 
After the candidates have been sorted into fronts $F_{1}$ to $F_{K}$ using nondominated sorting, the candidate that violates the constraint least will be reassigned to front $F_{K+1}$, the candidate with the second smallest violation to $F_{K+2}$, and so on. 
The concept is based on Deb et al.\ \cite{deb02}.  
Since the constraint violators are in the last fronts, they are less likely to be selected for the next generation.  

\subsubsection{Mutation} 
Since the aforementioned mutators do not take the data distribution into account and can potentially generate unlikely new candidates, we suggest a conditional mutator. 
It generates plausible feature values conditional on the values of the other features. For each input feature, we trained a transformation tree \cite{hothorn17} on $X^{obs}$, which is then used to sample values from the conditional distribution.
We mutate the feature in randomized order since a feature mutation now depends on the previous changes.\\[0.2cm]
How our proposed strategies for initialization and mutation affect MOC is later examined in a benchmark study (Sections~\ref{setup}~\&~\ref{results}).

\subsection{Evaluation Metric}
We use the popular hypervolume indicator (HV) \cite{zitzler98} to evaluate the quality of our estimated Pareto front, with reference point 
$\mathbf{s} = (\inf\limits_{y' \in Y'}|\hat{f}(\xv^*) - y'|, 1, p, 1)$,
representing the maximal values of the objectives.
We compute the HV always over the complete archive of evaluated solutions. 

\subsection{Tuning of Parameters}
We also use HV, when we tune MOC's control parameters -- population size, the probabilities for recombining and mutating a feature of a candidate -- 
with iterated F-racing \cite{lopez16}. 
Furthermore, we let iterated F-racing decide whether our proposed strategies for initialization and mutation of Section \ref{modification} are preferable. 
Tuning is performed on six binary classification datasets from OpenML \cite{vanschoren13} -- which were not used in the benchmark.
A summary of the tuning setup and results can be found in Table \ref{tab:psirace} in Appendix~\ref{ap:irace}. Iterated F-racing found both our initialization and mutation strategy to be advantageous.
The tuned parameters were used for the credit data application and the benchmark study. 

\section{Credit Data Application}
\label{example}

This section demonstrates the usefulness of MOC to explain the prediction of credit risk using the German credit dataset \cite{kaggle16}.
The dataset has 522 complete observations and nine features containing credit and customer information. Categories with few case numbers were combined. 
The binary target indicates whether a customer has a `good' or `bad' credit risk. 
We chose the first observation of the dataset as $\xv^*$ with the following feature values: 
	\begin{center}
	\begin{scriptsize}
		\begin{tabular}{ccccccccc}
			\toprule
			 Age & Sex & Job & Housing & Saving accounts & Checking account & Credit amount & Duration & Purpose  \\ 
			\midrule
		     22 & female &   2 & own & little & moderate & 5951 &  48 & radio/TV  \\ 
			\bottomrule
		\end{tabular}
	\end{scriptsize}
\end{center}
We tuned a support vector machine (with radial-basis (RBF) kernel) on the remaining data with the same tuning setup as for the benchmark (Appendix~\ref{ap:bench}). 
To obtain a single numerical outcome, only the predicted probability for the class `good' credit risk was returned. We obtained an accuracy of 0.64 for the model using two nested cross-validations (CV) (5-fold CV in outer and inner loop) and a predicted probability for `good' credit risk of $0.41$ for $\xv^*$.

We set the desired outcome interval to $Y' = [0.5, 1]$, which indicates a change to a `good' credit risk. 
We generated counterfactuals using MOC with the parameter setting selected by iterated F-racing. 
Candidates with a prediction below $0.5$ were penalized.  

\begin{figure}
\subfigure[Parallel plot]{
	\includegraphics[width=0.4\textwidth]{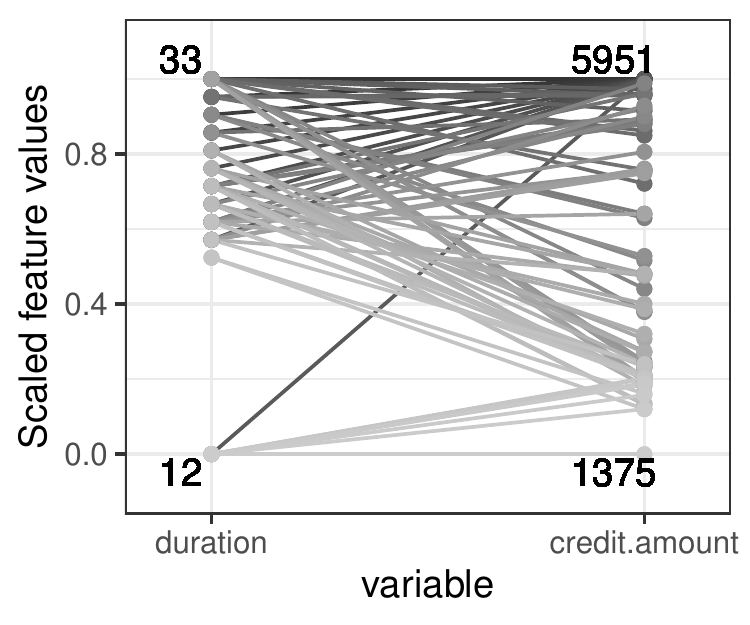}}
\subfigure[Response surface plot]{
	\includegraphics[width=0.54\textwidth]{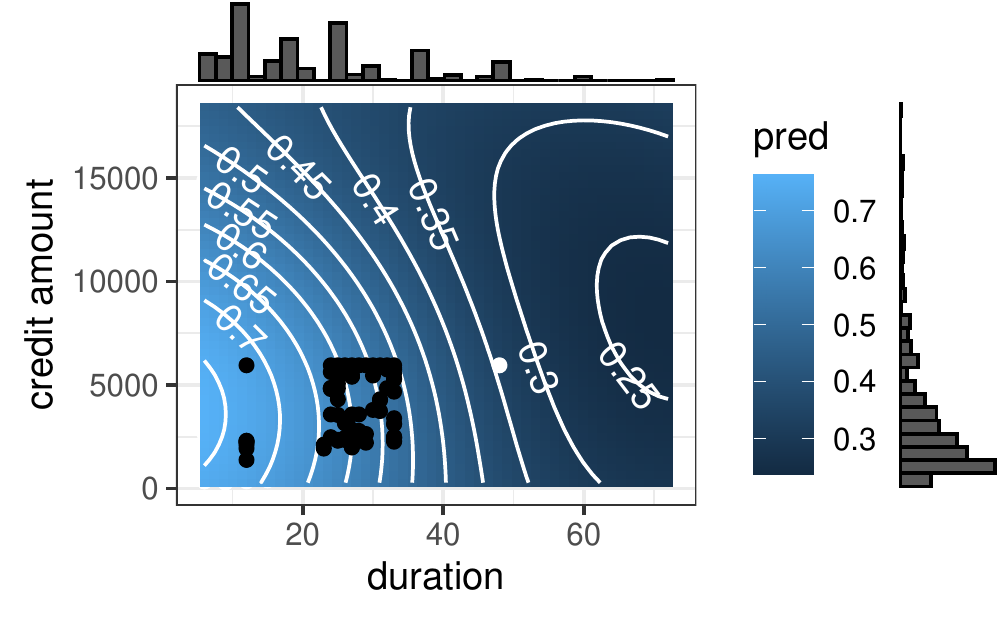}}
	\caption{Visualization of counterfactuals for the first data point $\xv^*$ of the credit dataset. \textbf{(a)} Feature values of the counterfactuals. Only changed features are shown. The given numbers indicate the minimum and maximum feature values of the counterfactuals. \textbf{(b)} Response surface plot for the model prediction along features duration and credit amount, holding other feature values constant at the value of $\xv^*$. Colors and contour lines indicate the predicted value. The white point is $\xv^*$ and the black points are the counterfactuals that only proposed changes in duration and/or credit amount. The histograms show the marginal distributions of the features in the observed dataset.}
\label{fig:cred}
\end{figure}

A total of 136 counterfactuals were found by MOC. In the following, we focus upon the 82 of them with predictions within $[0.5, 1]$.  
Credit \textit{duration} was changed for all counterfactuals, followed by \textit{credit amount} (86\%).
Since a user might not want to investigate all returned counterfactuals individually (in feature space), we provide a visual summary of the Pareto set in Figure \ref{fig:cred}, either as a parallel coordinate plot or a response surface plot\footnote{This is equivalent to a 2-D ICE-curve through $\xv^*$ \cite{goldstein15}. We refer to Section~\ref{modification} for a general definition of ICE curves.} along two features.
All counterfactuals had values equal to or smaller than the values of $\xv^*$ for \textit{duration} and  \textit{credit amount}.
The response surface plot illustrates why these feature changes were recommended.
The color gradient and contour lines indicate that either \textit{duration} or both  \textit{credit amount} and  \textit{duration} must be decreased to reach the desired outcome. 
Due to the fourth objective and the conditional mutator, we obtained counterfactuals in high density areas (indicated by histograms). 
Counterfactuals in the lower left corner seem to be in a less favorable region far from $\xv^*$, but they are close to the training data. 

\section{Experimental Setup}
\label{setup}
In this section, the performance of MOC is evaluated in a benchmark study for binary classification. 
The datasets are from the OpenML platform \cite{vanschoren13} and are briefly described in Table \ref{tab:databench}. 
We selected datasets with no missing values, with up to 3500 observations and a maximum of 40 features.
	\begin{table}[ht!]
	\begin{minipage}{.4\linewidth}
      \centering
        	\caption{Description of benchmark datasets. Legend: \textit{task:} OpenML task id; \textit{Obs:} Number of rows; \textit{Cont/Cat:} Number of continuous/categorical features.}
        	    \vspace{1.1cm}
        	    \scriptsize
        	\begin{tabular}{lllll}
		\toprule
		Task & Name & Obs & Cont & Cat \\ 
		\midrule
		3718 & boston & 506 &  12 &   1 \\ 
		3846 & cmc & 1473 &   2 &   7 \\ 
		145976 & diabetes & 768 &   8 &   0 \\ 
		9971 & ilpd & 583 &   9 &   1 \\ 
		3913 & kc2 & 522 &  21 &   0 \\ 
		3 & kr-vs-kp & 3196 &   0 &  36 \\ 
		3749 & no2 & 500 &   7 &   0 \\ 
		3918 & pc1 & 1109 &  21 &   0 \\ 
		3778 & plasma\_retinol & 315 &  10 &   3 \\ 
		145804 & tic-tac-toe & 958 &   0 &   9 \\ 
		\bottomrule
		\label{tab:databench}
	\end{tabular}
    \end{minipage} 
    \qquad
    \begin{minipage}{.57\linewidth}
      \caption{MOC's coverage rate of methods to be compared per dataset averaged over all models. The number of nondominated counterfactuals for each method are given in parentheses. Higher values of coverage indicate that MOC dominates the other method. The $^*$ indicates that the binomial test with $H_0: p < 0.5$ that a counterfactual is covered by MOC is significant at the $0.05$ level.}
      \centering
      \scriptsize
        \begin{tabular}{llrr}
		\toprule
		& DiCE & Recourse & Tweaking \\ 
		\midrule
		boston & 1* (36) & 0.92* (24) & 0.9* (10) \\ 
        cmc & 1* (17) &  & 0.75 (8) \\ 
        diabetes & 1* (64) & 0.45 (40) & 1 (3) \\ 
        ilpd & 1* (26) & 1* (37) & 0.83 (6) \\ 
        kc2 & 1* (53) & 0.31 (55) & 1 (2) \\ 
        kr-vs-kp & 1* (8) &  & 0.2 (10) \\ 
        no2 & 1* (58) & 0.5 (12) & 0.9* (10) \\ 
        pc1 & 1* (60) & 0.66* (38) &  \\ 
        plasma\_retinol & 1* (7) &  & 0.89* (9) \\ 
        tic-tac-toe & 1* (20) &  & 0.75 (8) \\ 
		\bottomrule
		\label{tab:cov}
	\end{tabular}
    \end{minipage}
\end{table}
We randomly selected ten observed data points per dataset as $\xv^*$ and excluded them from the training data. 
For each dataset, we tuned and trained the following models: logistic regression, random forest, xgboost, RBF support vector machine and a one-hidden-layer neural network. 
The tuning parameter set and the performance using nested resampling are in Table \ref{tab:perf} in Appendix~\ref{ap:bench}. 
Each model returned only the probability for one class.
The desired target for each $\xv^*$ was set to the opposite of the predicted class:
\begin{equation*} Y' = 
\begin{cases}
]0.5, 1] & \textnormal{if } \hat{f}(\xv^*) \le 0.5 \\
[0, 0.5] & \textnormal{else } 
\end{cases}.\end{equation*} 
The benchmark study aimed to answer two research questions:\nobreakpar
\begin{itemize}
	\setlength{\itemindent}{1em}
	\item[Q1)] How does MOC perform compared to other state-of-the-art methods for counterfactuals?
	\item[Q2)] How do our proposed strategies for initialization and mutation of Section~\ref{modification} influence the performance of MOC?   
\end{itemize}  
For the first one, we compared MOC -- once with and once without our proposed strategies for initialization and mutation -- with `DiCE' by Mothilal et al. \cite{mothilal19}, `Recourse' by Ustun et al. \cite{ustun19} and `Tweaking' by Tolomei et al. \cite{tolomei17}. 
We chose DiCE, Recourse and Tweaking because they are implemented in general open source code libraries.\footnote{Most other counterfactual methods are implemented for specific examples, but cannot be easily used for other datasets.}  
The methods are only applicable to certain models: DiCE can handle neural networks and logistic regressions, Recourse can handle logistic regressions and Tweaking can handle random forests. Since Recourse can only process binary and numerical features, we did not train logistic regression on cmc, tic-tac-toe, kr-vs-kp and plasma\_retinol. 
As a baseline, we selected the closest observed data point to $\xv^*$  (according to the Gower distance) that has a prediction equal to our desired outcome. Since this approach is part of the \textit{What-If Tool} \cite{pair19}, we call this approach `Whatif'.

The parameters of DiCE, Recourse and Tweaking were set to the default values recommended by the authors (Appendix~\ref{ap:paracf}). 
To allow for a fair comparison, we initialized MOC with the parameters of iterated F-racing which were tuned on other binary classification datasets (Appendix~\ref{ap:irace}). 
While MOC can potentially return several hundreds of counterfactuals, the other methods are designed to either return one or a few. 
We have therefore limited the maximum number of counterfactuals to ten for all approaches.\footnote{Note that this artificially penalizes our approach in the benchmark comparison.}
Tweaking and Whatif generated only one counterfactual by design. 
For MOC we reduced the number of counterfactuals by preferring the ones that achieved the target prediction $Y'$ and/or the highest HV contribution.

For all methods, only nondominated counterfactuals were considered for the evaluation. 
Since we are interested in a diverse set of counterfactuals, we evaluate the methods based on the size of their counterfactual set, its objective values, and the coverage rate derived from the coverage indicator by Zitzler and Thiele \cite{zitzler98}.
The coverage rate is the relative frequency with which counterfactuals of a method are dominated by MOC's counterfactuals for a certain model and $\xv^*$. 
A counterfactual covers another counterfactual if it dominates it, and it does not cover the other if both have the same objective values or the other has lower values in at least one objective. 
A coverage rate of 1 implies that for each generated counterfactual of a method MOC generated at least one dominating counterfactual.  
We only computed the coverage rate over counterfactuals that met the desired target $Y'$.

To answer the second research question, we compared the dominated HV over the generations of MOC with and without our proposed strategies for initialization and mutation. 
As a baseline, we used a random search approach that has the same population size (20) and number of generations (175) as MOC. 
In each generation, some feature values were uniformly sampled from their set of possible values derived from the observed data and $\xv^*$, while other features were set to the values of $\xv^*$.    
The HV for one generation was computed over the newly generated candidates combined with the candidates of the previous generations.

\section{Results}
\label{results}
\begin{figure*}[ht]
	\centering
	\subfigure[diabetes]{
		\includegraphics[width=0.49\linewidth]{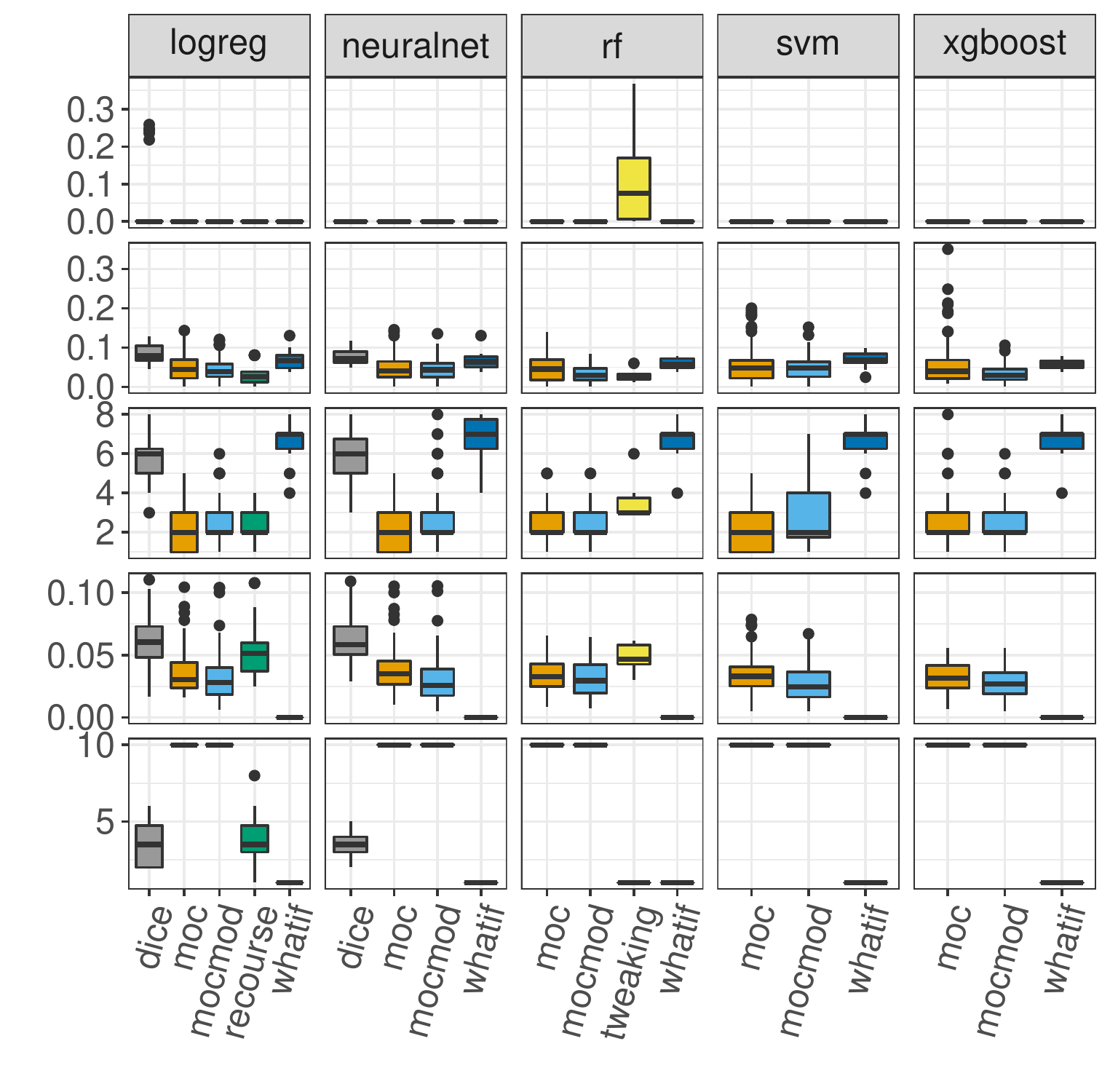}}
	\hspace{-0.38cm}
	\subfigure[no2]{
		\includegraphics[width=0.49\linewidth]{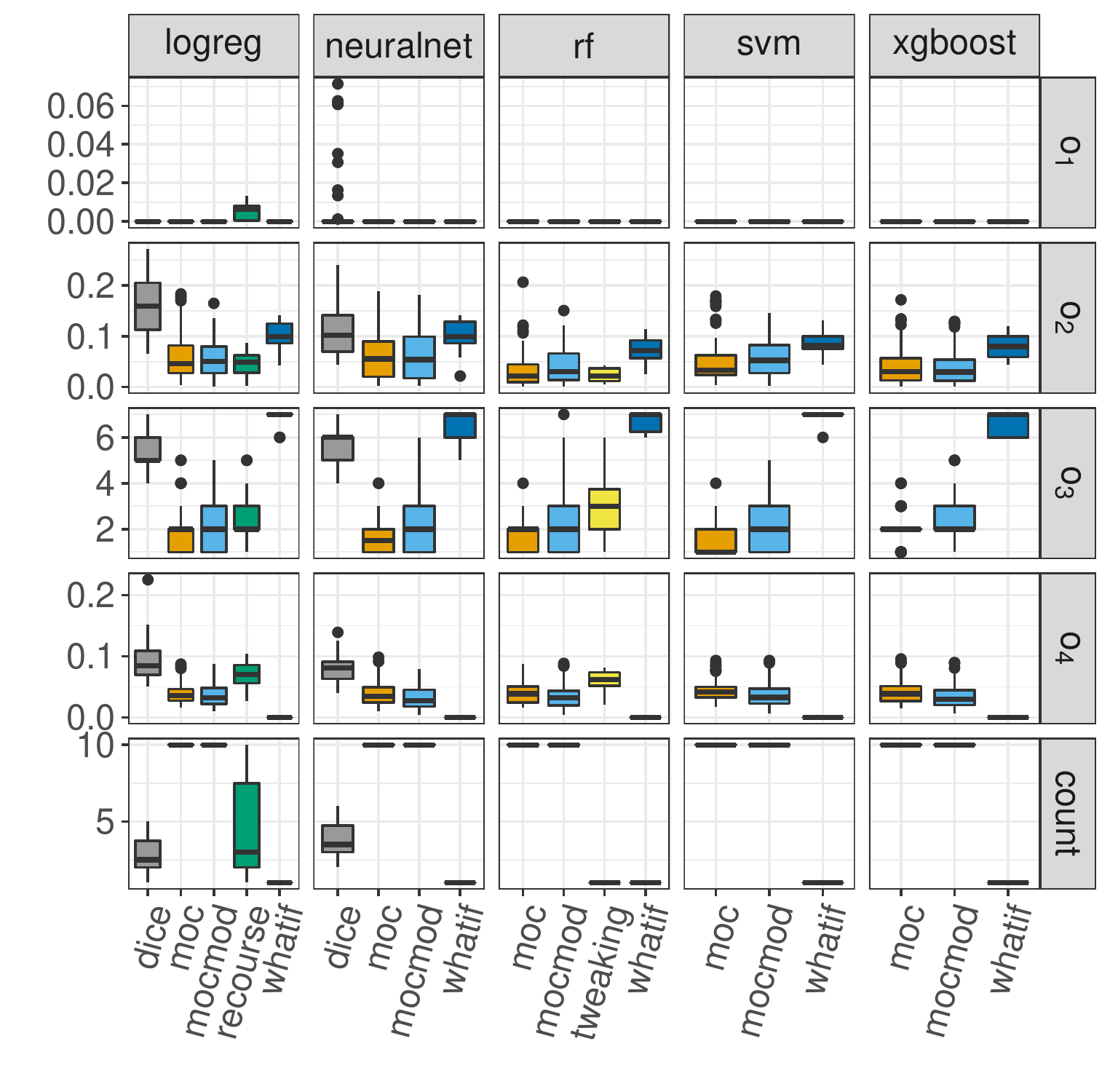}}
	\caption{Boxplots of the objective values and number of nondominated counterfactuals (\textit{count}) per model for MOC with our proposed strategies for initialization and mutation (\textit{mocmod}), MOC without these modifications, Whatif, DiCE, Recourse and Tweaking for the datasets diabetes and no2. Lower values are better except for \textit{count}.}
	\label{fig:boxshowed}
\end{figure*}
\subsection*{Q1) MOC vs. State-of-the-Art Counterfactual Methods}
Table \ref{tab:cov} shows the coverage rate of each method  (to be compared) by the tuned MOC per dataset. 
Some fields are empty because Recourse could not process features with more than two classes and Tweaking never achieved the desired outcome for pc1. 
MOC's counterfactuals dominated all counterfactuals of DiCE for all datasets. The same holds for Tweaking except for kr-vs-kp and tic-tac-toe because the counterfactuals of Tweaking had the same objective values as the ones of MOC. MOC's coverage rate of Recourse only exceeded 90\% for boston and ilpd since Recourse's counterfactuals often deviated less from $\xv^*$ (but performed worse in other objectives). 

Figure \ref{fig:boxshowed} compares MOC (with (\textit{mocmod}) and without (\textit{moc}) our proposed strategies for initialization and mutation) with the other methods for the datasets diabetes and no2 and for each model separately. 
The resulting boxplots for all other datasets are shown in Figures \ref{fig:otherboxes1} and \ref{fig:otherboxes2} in the Appendix. 
They agree with the results shown here. 
Compared to the other methods, both versions of MOC found the most nondominated solutions, which met the target and changed the least features. 
DiCE performed worse than MOC in all objectives.
Tweaking's counterfactuals were often closer to $\xv^*$, but they were further away from the nearest training data point and more features were changed. 
Tweaking's counterfactuals often did not reach the desired outcome because they stayed too close to $\xv^*$.
The MOC with our proposed modifications found counterfactuals closer to $\xv^*$ and the observed data, but required more feature changes compared to MOC without the modifications.

\subsection*{Q2) MOC Strategies for Initialization and Mutation}
\begin{figure}[ht]
	\centering
		\centerline{\includegraphics[width=0.6\columnwidth]{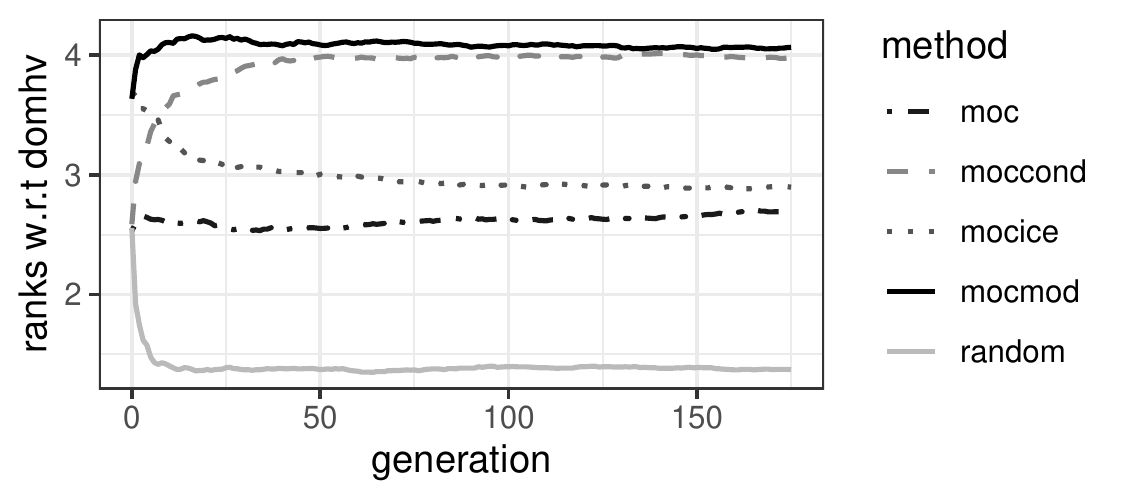}}
		\caption{Comparison of the ranks w.r.t. the dominated HV (\textit{domhv}) per generation averaged over all models and datasets. For each approach, the population size of each generation was 20. A higher HV and therefore a higher rank is better. Legend: \textit{moc}: MOC without our proposed modifications;   
			\textit{moccond}: MOC with the conditional mutator; \textit{mocice}: MOC with the ICE curve variance initialization; \textit{mocmod}: MOC with both modifications; \textit{random}: random search.}
		\label{fig:resall}
\end{figure}

Figure \ref{fig:resall} shows the ranks of the dominated HVs for MOC without modifications, for each modification of MOC and random search. Ranks were calculated per dataset, model, $\xv^*$ and generation, and were averaged over all datasets, models and $\xv^*$. We transformed HVs to ranks because the HVs are not comparable across $\xv^*$.
It can be seen that the MOC with our proposed modifications clearly outperforms the MOC without these modifications. 
The ranks of the initial population were higher when the ICE curve variance was used to initialize the candidates. 
The use of the conditional mutator led to higher dominated HVs over the generations. We received the best performance over the generations when both modifications were used. 
At each generation, all versions of MOC outperformed random search.
Figure~\ref{ap:hvdataset} in the Appendix shows the ranks over the generations for each dataset separately. They largely agree with the results shown here. 
The performance gains of MOC compared to random search were particularly evident for higher-dimensional datasets.  

\section{Conclusion and Outlook}
\label{conclusion}
In this paper, we introduced Multi-Objective Counterfactuals (MOC), which to the best of our knowledge is the first method to formalize the counterfactual search as a multi-objective optimization problem. 
Compared to state-of-the-art approaches, MOC  returns a diverse set of counterfactuals with different trade-offs between our proposed objectives. 
Furthermore, MOC is model-agnostic and suited for classification, regression and mixed feature spaces. 
We demonstrated the usefulness of MOC to explain a prediction on the German credit dataset and showed in a benchmark study that MOC finds more counterfactuals than other counterfactual methods that are closer to the training data and required fewer feature changes. Our proposed initialization strategy (based on ICE curve variances) and our conditional mutator resulted in higher performance in fewer evaluations and in counterfactuals that were closer to the data point we were interested in and to the observed data.

MOC has only been evaluated on binary classification, and only with respect to the dominated HV and the individual objectives.  
It is an open question how to let users select the counterfactuals that meet their -- a-priori unknown -- trade-off between the objectives. 
We leave these investigations to future research. 

\section{Electronic Submission}
\label{submission}
The complete code of the algorithm and the code to reproduce the experiments and results of this paper are available at \url{https://github.com/susanne-207/moc}. The implementation of MOC is based on our implementation of \cite{li13}, which we also used for \cite{binder19}.
We will provide an open source R library with our implementation of the method based on the \texttt{iml} package \cite{molnar18}.

\clearpage 
\bibliographystyle{splncs04}
\bibliography{literature.bib}

\begin{thebibliography}{10}
\providecommand{\url}[1]{\texttt{#1}}
\providecommand{\urlprefix}{URL }
\providecommand{\doi}[1]{https://doi.org/#1}

\bibitem{allaire19}
Allaire, J., Chollet, F.: keras: {R} {I}nterface to '{K}eras' (2019),
  \url{https://keras.rstudio.com}, {R} package version 2.3.0

\bibitem{avila06}
Avila, S.L., Kr{\"a}henb{\"u}hl, L., Sareni, B.: {A} {M}ulti-{N}iching
  {M}ulti-{O}bjective {G}enetic {A}lgorithm for {S}olving {C}omplex
  {M}ultimodal {P}roblems. In: {OIPE}. Sorrento, Italy (2006),
  \url{https://hal.archives-ouvertes.fr/hal-00398660}

\bibitem{binder19}
Binder, M., Moosbauer, J., Thomas, J., Bischl, B.: Multi-{O}bjective
  {H}yperparameter {T}uning and {F}eature {S}election using {F}ilter
  {E}nsembles (2019), accepted at {GECCO 2020}

\bibitem{bischl16}
Bischl, B., Lang, M., Kotthoff, L., Schiffner, J., Richter, J., Studerus, E.,
  Casalicchio, G., Jones, Z.M.: {mlr}: {M}achine {L}earning in {R}. Journal of
  Machine Learning Research  \textbf{17}(170), ~1--5 (2016),
  \url{http://jmlr.org/papers/v17/15-066.html}, {R} package version 2.17

\bibitem{breiman01}
Breiman, L.: Statistical {M}odeling: {T}he {T}wo {C}ultures. Statistical
  Science  \textbf{16}(3),  199--231 (08 2001). \doi{10.1214/ss/1009213726},
  \url{https://doi.org/10.1214/ss/1009213726}

\bibitem{deb95}
{Deb}, K., {Agarwal}, R.B.: Simulated {B}inary {C}rossover for {C}ontinuous
  {S}earch {S}pace. Complex Systems  \textbf{9},  115--148 (1995)

\bibitem{deb02}
Deb, K., Pratap, A., Agarwal, S., Meyarivan, T.: A {F}ast and {E}litist
  {M}ultiobjective {G}enetic {A}lgorithm: {NSGA-II}. IEEE Transactions on
  Evolutionary Computation  \textbf{6}(2),  182--197 (April 2002).
  \doi{10.1109/4235.996017}

\bibitem{dhurandhar19}
Dhurandhar, A., Pedapati, T., Balakrishnan, A., Chen, P., Shanmugam, K., Puri,
  R.: Model {A}gnostic {C}ontrastive {E}xplanations for {S}tructured {D}ata.
  CoRR  \textbf{abs/1906.00117} (2019), \url{http://arxiv.org/abs/1906.00117}

\bibitem{goldstein15}
Goldstein, A., Kapelner, A., Bleich, J., Pitkin, E.: Peeking {I}nside the
  {B}lack {B}ox: {V}isualizing {S}tatistical {L}earning {W}ith {P}lots of
  {I}ndividual {C}onditional {E}xpectation. Journal of Computational and
  Graphical Statistics  \textbf{24}(1),  44--65 (2015).
  \doi{10.1080/10618600.2014.907095},
  \url{https://doi.org/10.1080/10618600.2014.907095}

\bibitem{gower71}
Gower, J.C.: A {G}eneral {C}oefficient of {S}imilarity and {S}ome of its
  {P}roperties. Biometrics  \textbf{27}(4),  857--871 (1971)

\bibitem{grath18}
Grath, R.M., Costabello, L., Van, C.L., Sweeney, P., Kamiab, F., Shen, Z.,
  L{\'{e}}cu{\'{e}}, F.: Interpretable {C}redit {A}pplication {P}redictions
  {W}ith {C}ounterfactual {E}xplanations. CoRR (abs/1811.05245) (2018),
  \url{http://arxiv.org/abs/1811.05245}

\bibitem{greenwell18}
Greenwell, B.M., Boehmke, B.C., McCarthy, A.J.: A simple and effective
  model-based variable importance measure. arXiv preprint arXiv:1805.04755
  (2018)

\bibitem{kaggle16}
Hofmann, H.: {G}erman {C}redit {R}isk (2016),
  \url{https://www.kaggle.com/uciml/german-credit}, last accessed 25.01.2020

\bibitem{hothorn17}
Hothorn, T., Zeileis, A.: Transformation {F}orests (2017)

\bibitem{joshi19}
Joshi, S., Koyejo, O., Vijitbenjaronk, W., Kim, B., Ghosh, J.: Towards
  {R}ealistic {I}ndividual {R}ecourse and {A}ctionable {E}xplanations in
  black-box decision making systems. CoRR  \textbf{abs/1907.09615} (2019),
  \url{http://arxiv.org/abs/1907.09615}

\bibitem{karimi19}
Karimi, A., Barthe, G., Balle, B., Valera, I.: Model-{A}gnostic
  {C}ounterfactual {E}xplanations for {C}onsequential {D}ecisions. CoRR
  (abs/1905.11190) (2019), \url{http://arxiv.org/abs/1905.11190}

\bibitem{kingma14}
Kingma, D., Ba, J.: Adam: {A} {M}ethod for {S}tochastic {O}ptimization.
  International Conference on Learning Representations  (12 2014)

\bibitem{laugel17}
Laugel, T., Lesot, M.J., Marsala, C., Renard, X., Detyniecki, M.:
  Comparison-{B}ased {I}nverse {C}lassification for {I}nterpretability in
  {M}achine {L}earning. CoRR (abs/1712.08443) (2017),
  \url{http://arxiv.org/abs/1712.08443}

\bibitem{li13}
Li, R., Emmerich, M.T., Eggermont, J., B{\"{a}}ck, T., Sch{\"{u}}tz, M.,
  Dijkstra, J., Reiber, J.H.: Mixed {I}nteger {E}volution {S}trategies for
  {P}arameter {O}ptimization. Evolutionary Computation  \textbf{21}(1),  29--64
  (2013)

\bibitem{looveren19}
Looveren, A.V., Klaise, J.: Interpretable {C}ounterfactual {E}xplanations
  {G}uided by {P}rototypes. CoRR  \textbf{abs/1907.02584} (2019),
  \url{http://arxiv.org/abs/1907.02584}

\bibitem{lopez16}
L{\'o}pez-Ib{\'a}{\~n}ez, M., Dubois-Lacoste, J., C{\'a}ceres, L.P., Birattari,
  M., St{\"u}tzle, T.: The irace {P}ackage: {I}terated {R}acing for {A}utomatic
  {A}lgorithm {C}onfiguration. Operations Research Perspectives  \textbf{3},
  43 -- 58 (2016). \doi{https://doi.org/10.1016/j.orp.2016.09.002},
  \url{http://www.sciencedirect.com/science/article/pii/S2214716015300270}, {R}
  package version 3.4.1

\bibitem{lundberg17}
Lundberg, S.M., Lee, S.I.: A {U}nified {A}pproach to {I}nterpreting {M}odel
  {P}redictions. In: Advances in {N}eural {I}nformation {P}rocessing {S}ystems.
  pp. 4765--4774 (2017)

\bibitem{molnar18}
Molnar, C., Bischl, B., Casalicchio, G.: {i}ml: {A}n {R} package for
  {I}nterpretable {M}achine {L}earning. JOSS  \textbf{3}(26), ~786 (2018).
  \doi{10.21105/joss.00786},
  \url{http://joss.theoj.org/papers/10.21105/joss.00786}

\bibitem{mothilal19}
Mothilal, R.K., Sharma, A., Tan, C.: Explaining {M}achine {L}earning
  {C}lassifiers through {D}iverse {C}ounterfactual explanations. CoRR
  (abs/1905.07697) (2019), \url{http://arxiv.org/abs/1905.07697}

\bibitem{poyiadzi19}
Poyiadzi, R., Sokol, K., Santos-Rodriguez, R., Bie, T.D., Flach, P.: {FACE}:
  {F}easible and {A}ctionable {C}ounterfactual {E}xplanations (2019)

\bibitem{radulescu13}
Radulescu, A., L{\'o}pez-Ib{\'a}{\~{n}}ez, M., St{\"u}tzle, T.: {A}utomatically
  {I}mproving the {A}nytime {B}ehaviour of {M}ultiobjective {E}volutionary
  {A}lgorithms. In: Purshouse, R.C., Fleming, P.J., Fonseca, C.M., Greco, S.,
  Shaw, J. (eds.) Evolutionary Multi-Criterion Optimization. pp. 825--840.
  Springer Berlin Heidelberg, Berlin, Heidelberg (2013)

\bibitem{ribeiro16}
Ribeiro, M.T., Singh, S., Guestrin, C.: "{W}hy {S}hould {I} {T}rust {Y}ou?"
  {E}xplaining the {P}redictions of {A}ny {C}lassifier. In: Proceedings of the
  22nd ACM SIGKDD International Conference on Knowledge Discovery and Data
  Mining. pp. 1135--1144 (2016)

\bibitem{russell19}
Russell, C.: Efficient {S}earch for {D}iverse {C}oherent {E}xplanations. CoRR
  (abs/1901.04909) (2019), \url{http://arxiv.org/abs/1901.04909}

\bibitem{sharma19}
Sharma, S., Henderson, J., Ghosh, J.: {CERTIFAI:} {C}ounterfactual
  {E}xplanations for {R}obustness, {T}ransparency, {I}nterpretability, and
  {F}airness of {A}rtificial {I}ntelligence models. CoRR
  \textbf{abs/1905.07857} (2019), \url{http://arxiv.org/abs/1905.07857}

\bibitem{su17}
Su, J., Vargas, D.V., Sakurai, K.: One {P}ixel {A}ttack for {F}ooling {D}eep
  {N}eural {N}etworks. IEEE Transactions on Evolutionary Computation
  \textbf{23},  828--841 (2017)

\bibitem{syswerda89}
Syswerda, G.: Uniform {C}rossover in {G}enetic {A}lgorithms. In: Proceedings of
  the 3rd International Conference on Genetic Algorithms. p. 2–9. Morgan
  Kaufmann Publishers Inc., San Francisco, CA, USA (1989)

\bibitem{tolomei17}
Tolomei, G., Silvestri, F., Haines, A., Lalmas, M.: Interpretable {P}redictions
  of {T}ree-based {E}nsembles via {A}ctionable {F}eature {T}weaking. In:
  Proceedings of the 23rd ACM SIGKDD International Conference on Knowledge
  Discovery and Data Mining. pp. 465--474. KDD '17, ACM, New York, NY, USA
  (2017). \doi{10.1145/3097983.3098039},
  \url{http://doi.acm.org/10.1145/3097983.3098039}

\bibitem{ustun19}
Ustun, B., Spangher, A., Liu, Y.: Actionable {R}ecourse in {L}inear
  {C}lassification. In: Proceedings of the Conference on Fairness,
  Accountability, and Transparency. pp. 10--19. FAT* '19, ACM, New York, NY,
  USA (2019). \doi{10.1145/3287560.3287566},
  \url{http://doi.acm.org/10.1145/3287560.3287566}

\bibitem{vanschoren13}
Vanschoren, J., van Rijn, J.N., Bischl, B., Torgo, L.: {OpenML}: {N}etworked
  {S}cience in {M}achine {L}earning. SIGKDD Explorations  \textbf{15}(2),
  49--60 (2013). \doi{10.1145/2641190.2641198},
  \url{http://doi.acm.org/10.1145/2641190.2641198}

\bibitem{wachter17}
Wachter, S., Mittelstadt, B.D., Russell, C.: Counterfactual {E}xplanations
  without {O}pening the {B}lack {B}ox: {A}utomated {D}ecisions and the {GDPR}.
  CoRR (abs/1711.00399) (2017), \url{http://arxiv.org/abs/1711.00399}

\bibitem{pair19}
Wexler, J., Pushkarna, M., Bolukbasi, T., Wattenberg, M., Vi{\'{e}}gas, F.B.,
  Wilson, J.: The {W}hat- {I}f {T}ool: {I}nteractive {P}robing of {M}achine
  {L}earning {M}odels. CoRR  \textbf{abs/1907.04135} (2019),
  \url{http://arxiv.org/abs/1907.04135}

\bibitem{white19}
White, A., d'Avila Garcez, A.: Measurable {C}ounterfactual {L}ocal
  {E}xplanations for {A}ny {C}lassifier (2019)

\bibitem{zitzler98}
Zitzler, E., Thiele, L.: Multiobjective {O}ptimization {U}sing {E}volutionary
  {A}lgorithms --- a {C}omparative {C}ase {C}tudy. In: Eiben, A.E., B{\"a}ck,
  T., Schoenauer, M., Schwefel, H.P. (eds.) Parallel Problem Solving from
  Nature --- PPSN V. pp. 292--301. Springer Berlin Heidelberg, Berlin,
  Heidelberg (1998)

\end{thebibliography}

\clearpage
\appendix
\section{Illustration of MOC's Benefits}
\label{ap:examples}
This section illustrates the benefits of having a \emph{diverse set} of counterfactuals using the diabetes dataset of the benchmark study (Section~\ref{setup}).
We will compare the counterfactuals returned by MOC with the ones of Recourse \cite{ustun19} and Tweaking \cite{tolomei17}. Due to space constraints, we only show the six counterfactuals of MOC with the highest HV contribution for both examples.

Table~\ref{tab:cfdiab1} contrasts MOC's counterfactuals with the three counterfactuals of Recourse for the prediction of observation 741. 
A logistic regression predicted a probability of having diabetes of 0.89 for this observation. The desired target is a prediction of less than 0.5, which indicates having no diabetes. 
\begin{table}[ht]
	\caption{Counterfactuals and corresponding objective values of MOC and Recourse for the prediction of a logistic regression for observation 741 of the diabetes dataset. Shaded fields indicate values that differ from the value of observation 741 in brackets.}
	\label{tab:cfdiab1}
	\centering
	\scriptsize
	\begin{tabular}{lrrrrrrrrrr}
		\toprule
		Feature ($\xv^*$) & MOC$_1$ & MOC$_2$ & MOC$_3$ & MOC$_4$ & MOC$_5$ & MOC$_6$ & Recourse$_1$ & Recourse$_2$ & Recourse$_3$\\ 
		\midrule
	 preg (11) & 11.00 & \cellcolor{gray!25}6.35 & 11.00 & 11.00 & 11.00 & \cellcolor{gray!25}6.35 & 11.00 & 11.00 & \cellcolor{gray!25}10.92 \\ 
  plas (120) & \cellcolor{gray!25}27.78 & \cellcolor{gray!25}3.29 & \cellcolor{gray!25}79.75 & \cellcolor{gray!25}94.85 & \cellcolor{gray!25}79.75 & \cellcolor{gray!25}3.18 & \cellcolor{gray!25}57.00 & \cellcolor{gray!25}57.00 & \cellcolor{gray!25}57.00 \\ 
  pres (80) & 80.00 & 80.00 & 80.00 & 80.00 & 80.00 & 80.00 & 80.00 & 80.00 & 80.00 \\ 
  skin (37) & 37.00 & 37.00 & 37.00 & 37.00 & 37.00 & 37.00 & 37.00 & \cellcolor{gray!25}36.81 & 37.00 \\ 
  insu (150) & 150.00 & 150.00 & \cellcolor{gray!25}17.13 & 150.00 & \cellcolor{gray!25}40.61 & 150.00 & 150.00 & 150.00 & 150.00 \\ 
  mass (42.3) & 42.30 & 42.30 & \cellcolor{gray!25}29.17 & \cellcolor{gray!25}15.36 & \cellcolor{gray!25}29.17 & 42.30 & 42.30 & 42.30 & 42.30 \\ 
  pedi (0.78) & 0.78 & 0.78 & \cellcolor{gray!25}0.31 & 0.78 & \cellcolor{gray!25}0.17 & 0.78 & 0.78 & 0.78 & 0.78 \\ 
  age (48) & 48.00 & \cellcolor{gray!25}41.61 & \cellcolor{gray!25}44.42 & 48.00 & 48.00 & 48.00 & \cellcolor{gray!25}28.36 & \cellcolor{gray!25}28.36 & \cellcolor{gray!25}28.36 \\ 
  \midrule
  $o_1$  & 0.00 & 0.00 & 0.00 & 0.00 & 0.00 & 0.00 & 0.00 & 0.00 & 0.00 \\ 
  $o_2$  & 0.06 & 0.12 & 0.10 & 0.07 & 0.10 & 0.11 & 0.08 & 0.08 & 0.08 \\ 
  $o_3$  & 1.00 & 3.00 & 5.00 & 2.00 & 4.00 & 2.00 & 2.00 & 3.00 & 3.00 \\ 
  $o_4$  & 0.10 & 0.05 & 0.03 & 0.07 & 0.04 & 0.07 & 0.09 & 0.09 & 0.09 \\ 
		\bottomrule
	\end{tabular}
\end{table}
All counterfactuals of Recourse suggest the same reduction in \textit{age} and plasma concentration (\textit{plas}), with two counterfactuals additionally suggesting a minimal reduction in the number of pregnancies (\textit{preg}) or the skin fold thickness (\textit{skin}).\footnote{By reclassifying \textit{age} and \textit{preg} as integers (instead of decimals), integer changes would be recommended by MOC, Recourse and Tweaking.} 
Apart from that a reduction in \textit{age} or  \textit{preg} is impossible, they do not offer many options for users. Instead, MOC returned a larger set of counterfactuals that provide more options for actionable user responses and are closer to the observed data than Recourse's counterfactuals (\textit{$o_4$}).
Counterfactual MOC$_1$ has overall lower objective values than all counterfactuals of Recourse. MOC$_3$ suggested changes to five features so that it is especially close to the nearest training data point (\textit{$o_4$}).  

Table~\ref{tab:cfdiab2} compares the set of counterfactuals found by MOC with the single counterfactual found by Tweaking for the prediction of observation 268. 
A random forest classifier predicted a probability of having diabetes of 0.62 for this observation. Again, the desired target is a prediction of less than 0.5.
\begin{table}[ht]
	\caption{Counterfactuals and corresponding objective values given by MOC and Tweaking for the prediction of a random forest for observation 268 of the cmc dataset. Shaded fields indicate values that differ from the value of observation 268 in brackets.}
	\label{tab:cfdiab2}
	\centering
	\scriptsize
	\begin{tabular}{lrrrrrrr}
		\toprule
		Feature ($\xv^*$) & MOC$_1$ & MOC$_2$ & MOC$_3$ & MOC$_4$ & MOC$_5$ & MOC$_6$  & Tweaking$_1$ \\ 
		\midrule
preg (2) & 2.00 & 2.00 & 2.00 & 2.00 & 2.00 & 2.00 & \cellcolor{gray!25}1.53 \\ 
  plas (128) & \cellcolor{gray!25}121.50 & \cellcolor{gray!25}90.21 & \cellcolor{gray!25}126.83 & 128.00 & \cellcolor{gray!25}88.44 & \cellcolor{gray!25}120.64 & \cellcolor{gray!25}119.71 \\ 
  pres (64) & 64.00 & 64.00 & 64.00 & 64.00 & 64.00 & 64.00 & 64.00 \\ 
  skin (42) & 42.00 & 42.00 & 42.00 & 42.00 & 42.00 & 42.00 & 42.00 \\ 
  insu (0) & 0.00 & 0.00 & 0.00 & 0.00 & 0.00 & \cellcolor{gray!25}90.93 & 0.00 \\ 
  mass (40) & 40.00 & 40.00 & 40.00 & 40.00 & 40.00 & 40.00 & 40.00 \\ 
  pedi (1.1) & 1.10 & \cellcolor{gray!25}0.48 & 1.10 & \cellcolor{gray!25}0.17 & \cellcolor{gray!25}0.46 & 1.10 & 1.10 \\ 
  age (24) & 24.00 & 24.00 & 24.00 & 24.00 & \cellcolor{gray!25}25.85 & 24.00 & \cellcolor{gray!25}28.29 \\ 
  		\midrule
  $o_1$  & 0.00 & 0.00 & 0.00 & 0.00 & 0.00 & 0.00 & 0.00 \\ 
  $o_2$ & 0.00 & 0.06 & 0.00 & 0.05 & 0.06 & 0.02 & 0.02 \\ 
  $o_3$  & 1.00 & 2.00 & 1.00 & 1.00 & 3.00 & 2.00 & 3.00 \\ 
  $o_4$ & 0.05 & 0.02 & 0.05 & 0.04 & 0.01 & 0.03 & 0.06 \\ 
		\bottomrule
	\end{tabular}
\end{table}
Tweaking suggested reducing the number of children and plasma glucose concentration (\textit{plas}) while increasing the \textit{age} so that the probability of diabetes decreases. This is contradictory and not plausible. In contrast, MOC's counterfactuals suggest various strategies, e.g., only a decrease of \textit{plas}, which is easier to realize. In addition, MOC$_1$, MOC$_3$ and MOC$_6$ dominate the counterfactual of Tweaking. 
Since five of six counterfactuals suggest changes to \textit{plas}, the user may have more confidence that \textit{plas} is an important lever to achieve the desired outcome.

\section{Iterated F-racing}
\label{ap:irace}
We used iterated F-racing (irace) \cite{lopez16} to tune the parameters of MOC for binary classification. The parameters and considered ranges are given in Table \ref{tab:psirace}. 
The number of generations was not part of the parameter set because it would be always tuned to the upper bound. 
Instead, the number of generations was determined after the other parameters were tuned with irace. 
Irace was initialized with a maximum budget of 3000 evaluations equal to 3000 runs of MOC. 
\begin{table}[ht]
	\centering
	\caption{Parameter space investigated with iterated F-racing, as well as the resulting optimized configuration (\textit{Result}).} 
	\begin{scriptsize}
		\begin{tabular}{llll}
			\toprule
			Name & Description& Range & Result \\
			\midrule
			$M$ & Population size &  [20, 100] & 20 \\
			initialization & Initialization strategy & [Random, ICE curve] & ICE curve \\
			conditional & Whether to use the  & [TRUE, FALSE] & TRUE \\
			& 	conditional mutator & & \\
			p.rec & Probability a pair of & [0.3, 1] & 0.57\\
			& parents is chosen to  & & \\
			& recombine & & \\
			p.rec.gen & Probability a feature & [0.3, 1] & 0.85 \\
			& is recombined &   \\
			p.rec.use.orig & Probability the indicator & [0.3, 1] & 0.88 \\
			& for feature changes is \\
			& recombined  \\
			p.mut & Probability a child is &  [0.05, 0.8] & 0.79 \\
			& chosen to be mutated  \\
			p.mut.gen & Probability one  &  [0.05, 0.8] & 0.56 \\
			& feature is mutated  \\
			p.mut.use.orig & Probability indicator & [0.05, 0.5] & 0.32 \\
			& for a feature change is \\
			& flipped &  \\
			\bottomrule
		\end{tabular}
	\end{scriptsize}
	\label{tab:psirace}
\end{table}  
In every step, irace randomly selected one of 300 instances.  
Each instance consisted of a trained model, a randomly selected data point from the observed data as $\xv^*$ and a desired outcome.
The desired target for each $\xv^*$ was the opposite of the predicted class: 
$$
 Y' =
    \begin{cases}
    ]0.5, 1] & \textnormal{if } \hat{f}(\xv^*) \le 0.5 \\
    [0, 0.5] & \textnormal{else }
    \end{cases}.
$$  
The trained model was either logistic regression, random forest, xgboost, RBF support vector machine or a two-hidden-layer neural network. 
Each model estimated only the probability for one class. 
The models were trained on datasets obtained from the OpenML platform \cite{vanschoren13} (without the sampled $\xv^*$) and are briefly described in Table \ref{tab:datairace}. 
While these datasets were not used in the benchmark study (Section~\ref{setup}), the same preprocessing steps were conducted and the models were tuned with the same setup (see Section~\ref{ap:bench} for details).

In each step of irace, parameter configurations were evaluated by running MOC on the same selected instance. 
MOC stopped after evaluating 8000 candidates with Eq.~(\ref{eq:moo}), which should be enough to ensure convergence of the HV in most cases. 
The integral of the first order spline approximation of the dominated HV over the evaluations was the performance criterion as recommended by \cite{radulescu13}. The integral takes into account not only the extent but also the rate of convergence of the dominated HV.  
A Friedman test was used to discard less promising configurations. 
The first Friedman test was conducted after initial configurations were evaluated on 15 instances; afterward, the test was conducted after evaluating the remaining configurations on a single instance to accelerate the exclusion process.
The best configuration returned is given in Table \ref{tab:psirace}.   

To obtain a default parameter for the number of generations for the benchmark study, we determined for the 300 instances after how many generations of the tuned MOC the dominated HV has not increased for 10 generations. 
We chose the maximum of 175 generations as a default for the study. 

\begin{table}[ht]
	\begin{minipage}{.47\linewidth}
		\caption{Tuning search space per model. The hyperparameters \textit{ntrees} and \textit{nrounds} were log-transformed.} 
		\vspace{0.8cm}
		\scriptsize
		\centering
		\begin{tabular}{rll}
			\toprule
			Model & Hyperparameter & Range \\
			\midrule
			randomforest & ntrees & [0, 1000] \\
			xgboost & nrounds &  [0, 1000] \\
			svm & cost & [0.01, 1] \\
			logreg & lr & [0.0005, 0.1] \\
			neuralnet & lr &  [0.0005, 0.1] \\
			& layer\_size & [1, 6] \\
			\bottomrule
		\end{tabular}
		\label{tab:modeltunset}
	\end{minipage} 
	\qquad
	\begin{minipage}{.47\linewidth}  
		\caption[Datasets]{Description of datasets for tuning with iterated F-racing. 
			Legend: \textit{Task:} OpenML task id; \textit{Obs:} Number of rows; \textit{Cont/Cat:} Number of continuous/categorical features.} 
		\scriptsize
		\centering
		\begin{tabular}{rlllllcc}
			\toprule
			Task & Name & Obs & Cont & Cat \\ 
			\midrule
			3818 & tae & 151 &   3 &   2  \\ 
			3917 & kc1 & 2109 &   21 & 0  \\
			52945 & breastTumor & 277 &  0 &   6  \\ 	
			3483 & mammography & 11183 &   6 &   0  \\ 
			3822 & nursery & 12960 &   0 &   8 \\ 
			3586 & abalone & 4177 &    7 &   1 \\
			\bottomrule
		\end{tabular}
		\label{tab:datairace}
	\end{minipage}
\end{table} 

\section{Model Hyperparameters for the Benchmark Study}
\label{ap:bench}
We used random search (with 200 iterations for neural networks and 100 iterations for all other models) and 5-fold CV (with misclassification error as performance measure) to tune the hyperparameters of the models on the training data. 
The tuning search space was the same as for iterated F-racing and is shown in Table \ref{tab:modeltunset}.
Numerical features were scaled (standardization (Z-score) for random forest, min-max-scaling (0-1-range) for all other models) and categorical features were one-hot encoded.  
For neural network and logistic regression, ADAM \cite{kingma14} was the optimizer, the batch size was 32 with a 1/3 validation split and early stopping was conducted after 5 patience steps. 
Logistic regression needed these configurations because we constructed the model as a zero-hidden-layer neural network.
For all other hyperparameters of the models, we chose the default values of the \texttt{mlr} \cite{bischl16} and \texttt{keras} \cite{allaire19} R packages. 
Table \ref{tab:perf} shows the accuracies of the trained models using nested resampling (5-fold CV in outer and inner loop).
\begin{table}[ht]
	\centering
	\caption{Accuracy using nested resampling per benchmark dataset and model. Legend: \textit{Name:} OpenML task name; \textit{rf:} random forest. Logistic regression (\textit{logreg}) was only trained on datasets with numerical or binary features.}
	\label{tab:perf}
	\begingroup\scriptsize
	\begin{tabular}{lrrrrr}
		\toprule
		Name & rf & xgboost & svm & logreg & neuralnet \\ 
		\midrule
	boston & 0.90 & 0.89 & 0.87 & 0.86 & 0.87 \\ 
	cmc & 0.70 & 0.72 & 0.67 &  & 0.68 \\ 
	diabetes & 0.76 & 0.74 & 0.75 & 0.63 & 0.68 \\ 
	ilpd & 0.69 & 0.67 & 0.65 & 0.53 & 0.58 \\ 
	kc2 & 0.81 & 0.80 & 0.79 & 0.75 & 0.72 \\ 
	kr-vs-kp & 0.99 & 0.99 & 0.97 &  & 0.99 \\ 
	no2 & 0.63 & 0.59 & 0.58 & 0.55 & 0.54 \\ 
	pc1 & 0.93 & 0.93 & 0.91 & 0.91 & 0.88 \\ 
	plasma\_retinol & 0.53 & 0.52 & 0.58 &  & 0.55 \\ 
	tic-tac-toe & 0.99 & 0.99 & 0.98 &  & 0.97 \\ 
		\bottomrule
	\end{tabular}
	\endgroup
\end{table}

\section{Control Parameters of Counterfactual Methods}
\label{ap:paracf}
For Tweaking \cite{tolomei17}, we only changed $\epsilon$, a positive threshold that limits the tweaking of each feature. It was set to $0.5$ because it obtained better results for the authors on their data example on Ad Quality in comparison to the default value $0.1$. We used the R implementation of Tweaking on Github: \url{https://github.com/katokohaku/featureTweakR} (commit \texttt{6f3e614}). 
For Recourse \cite{ustun19}, we left all parameters at their default settings. We used the Python implementation of Recourse on Github: \url{https://github.com/ustunb/actionable-recourse} (commit \texttt{aaae8fa}).
For DiCE \cite{mothilal19}, we used the `DiverseCF' version proposed by the authors \cite{mothilal19} and left the control parameters at their defaults.
We used the inverse mean absolute deviation for the feature weights.
For datasets where the mean absolute deviation of a feature was zero, we set the feature weight to 10.
We used the Python implementation of DiCE available on Github: \url{https://github.com/microsoft/DiCE} (commit \texttt{fed9d27}).

\vspace{-2cm}
\begin{figure}[ht]
	\centering
	\subfigure[boston]{
		\includegraphics[width=0.49\linewidth]{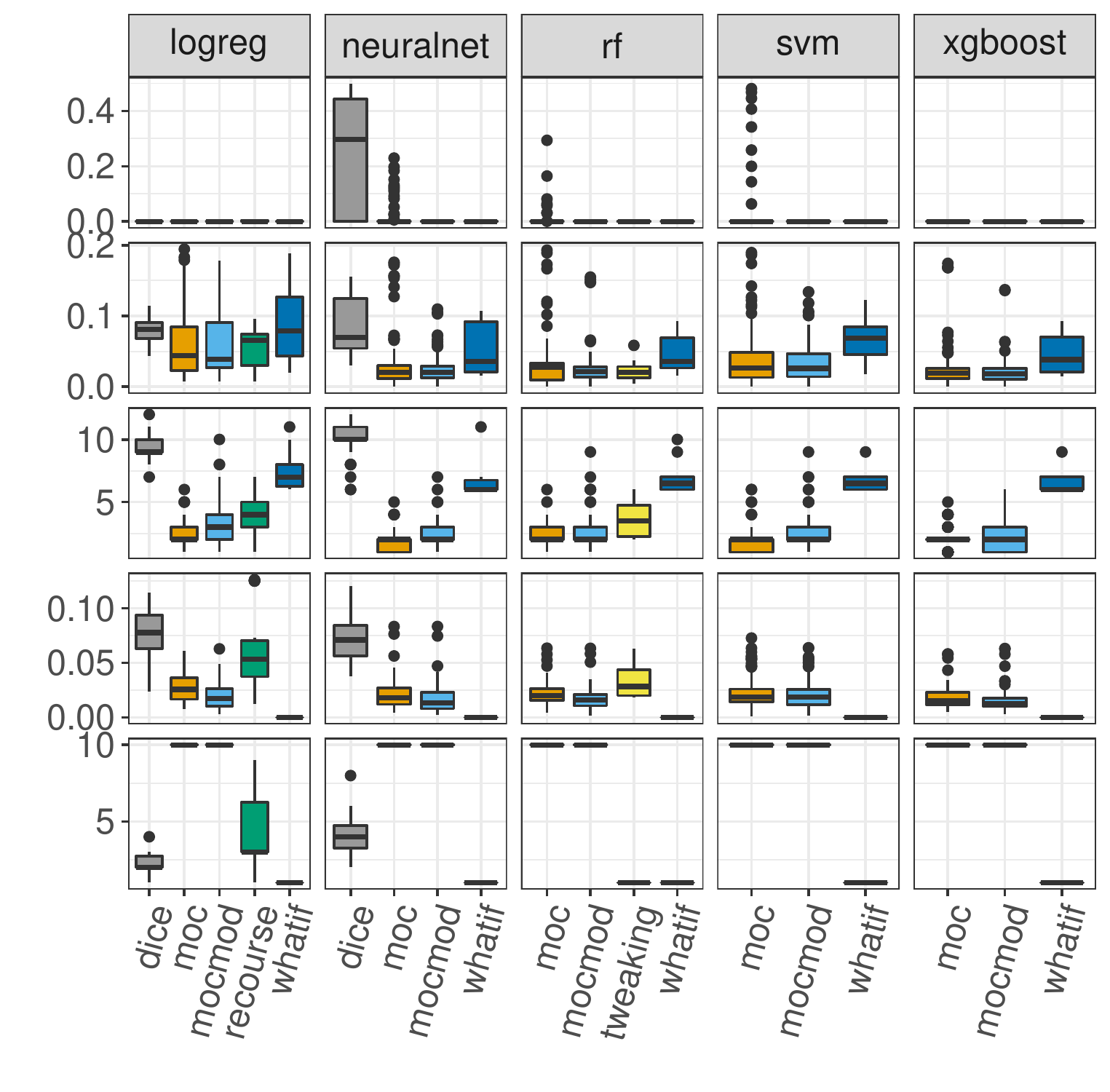}}
	\hspace{-0.38cm}
	\subfigure[pc1]{
		\includegraphics[width=0.49\linewidth]{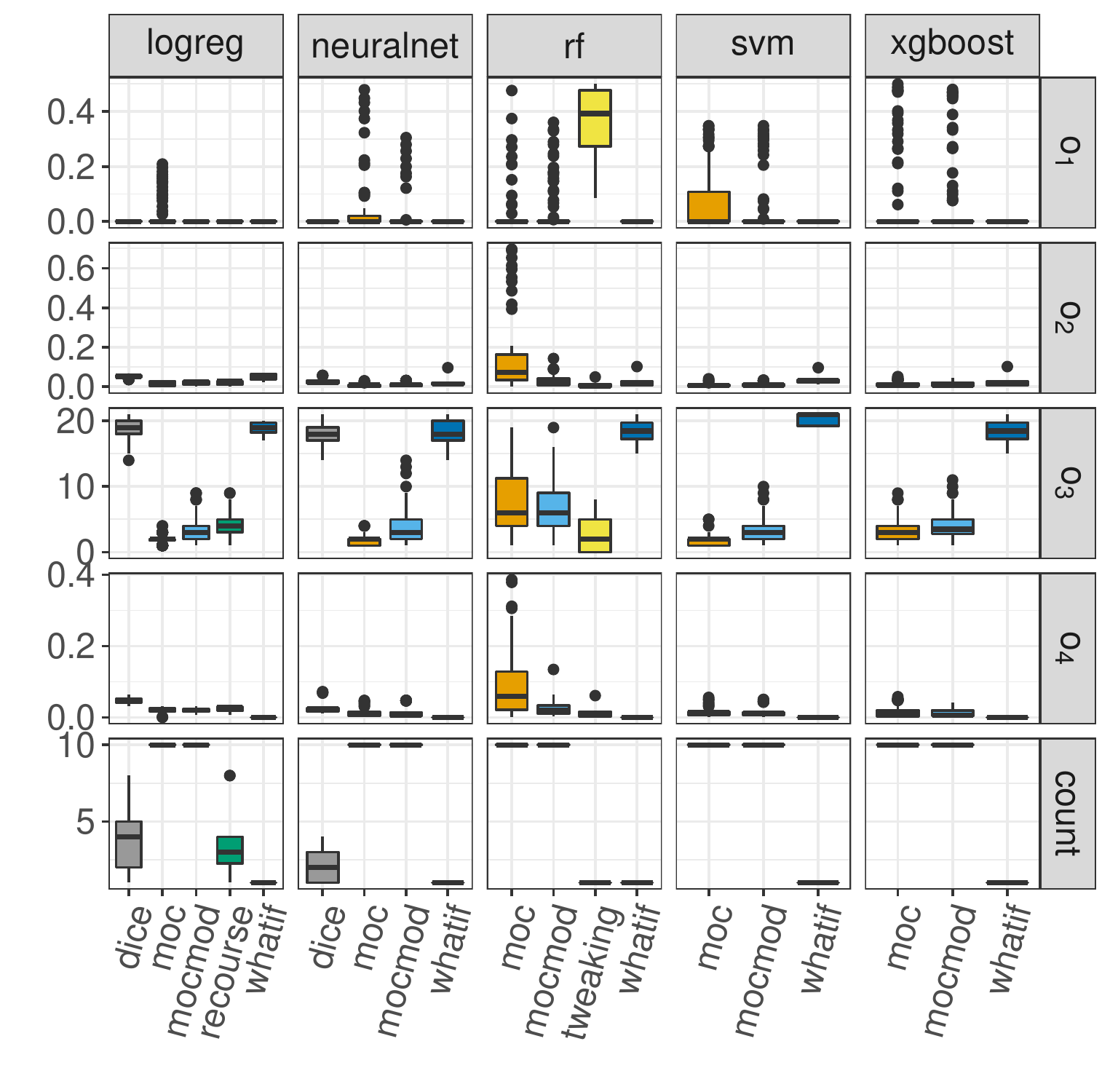}}
	\par\medskip
	\subfigure[ilpd]{
		\includegraphics[width=0.49\linewidth]{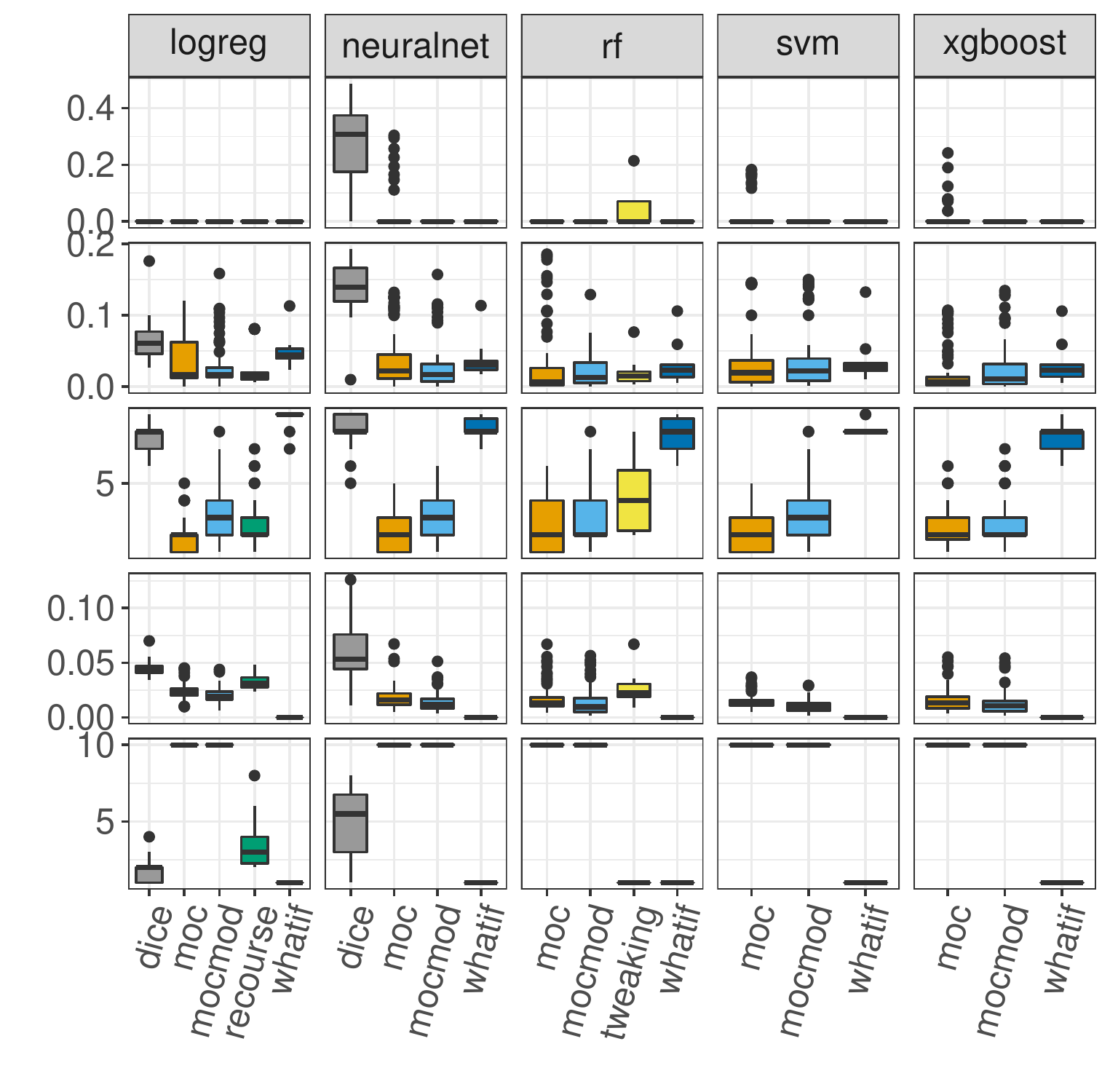}}
	\hspace{-0.38cm}
	\subfigure[kc2]{
		\includegraphics[width=0.49\linewidth]{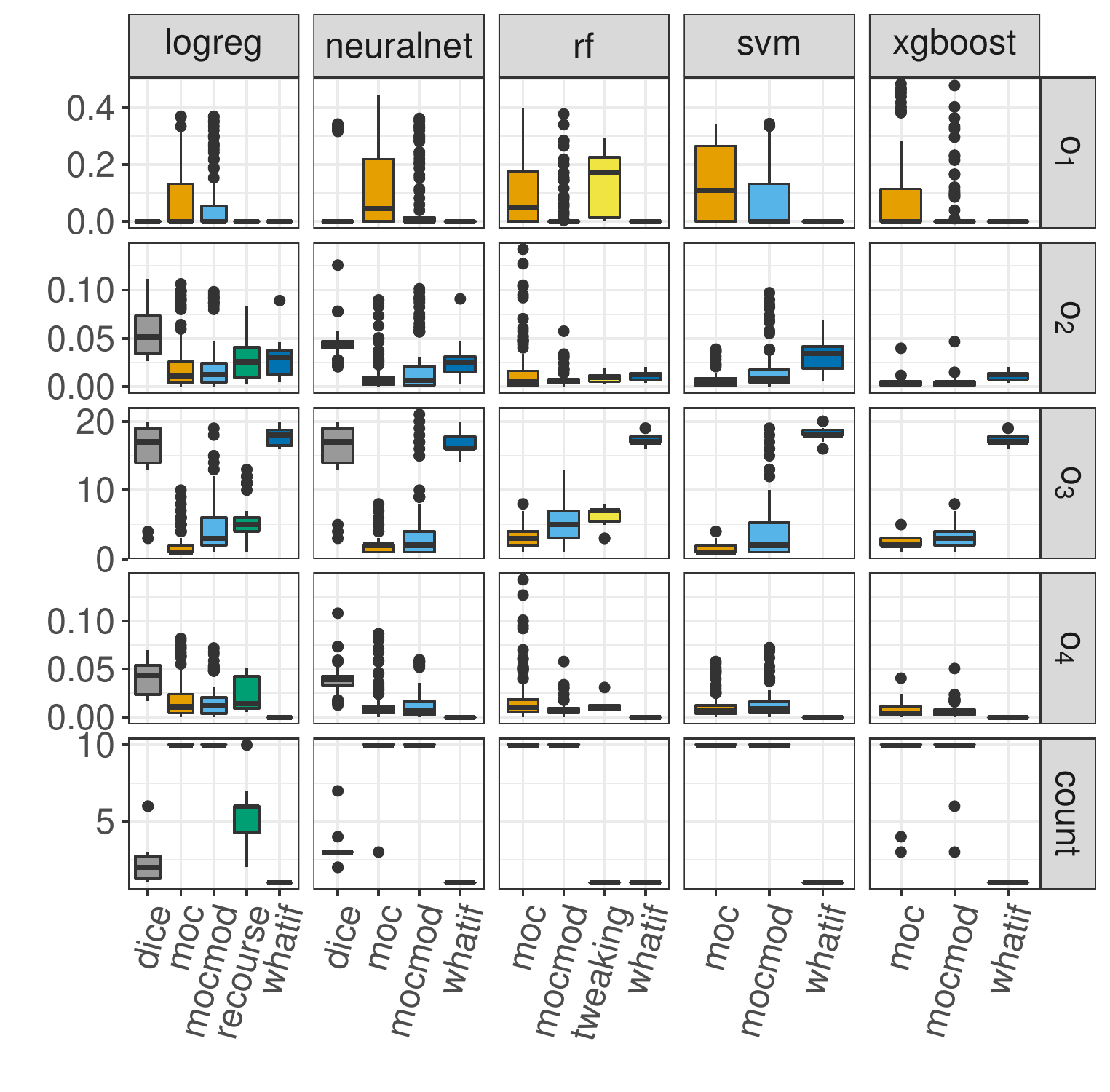}}
	\caption{Boxplots of the objective values and number of nondominated counterfactuals (\textit{count}) per dataset and model for MOC with our proposed strategies for initialization and mutation (\textit{mocmod}), MOC without these modifications, Whatif, DiCE, Recourse and Tweaking. Lower values are better except for \textit{count}.}
	\label{fig:otherboxes1}
\end{figure}

\begin{figure}[ht]
	\centering
	\subfigure[kr-vs-kp]{
		\includegraphics[width=0.49\linewidth]{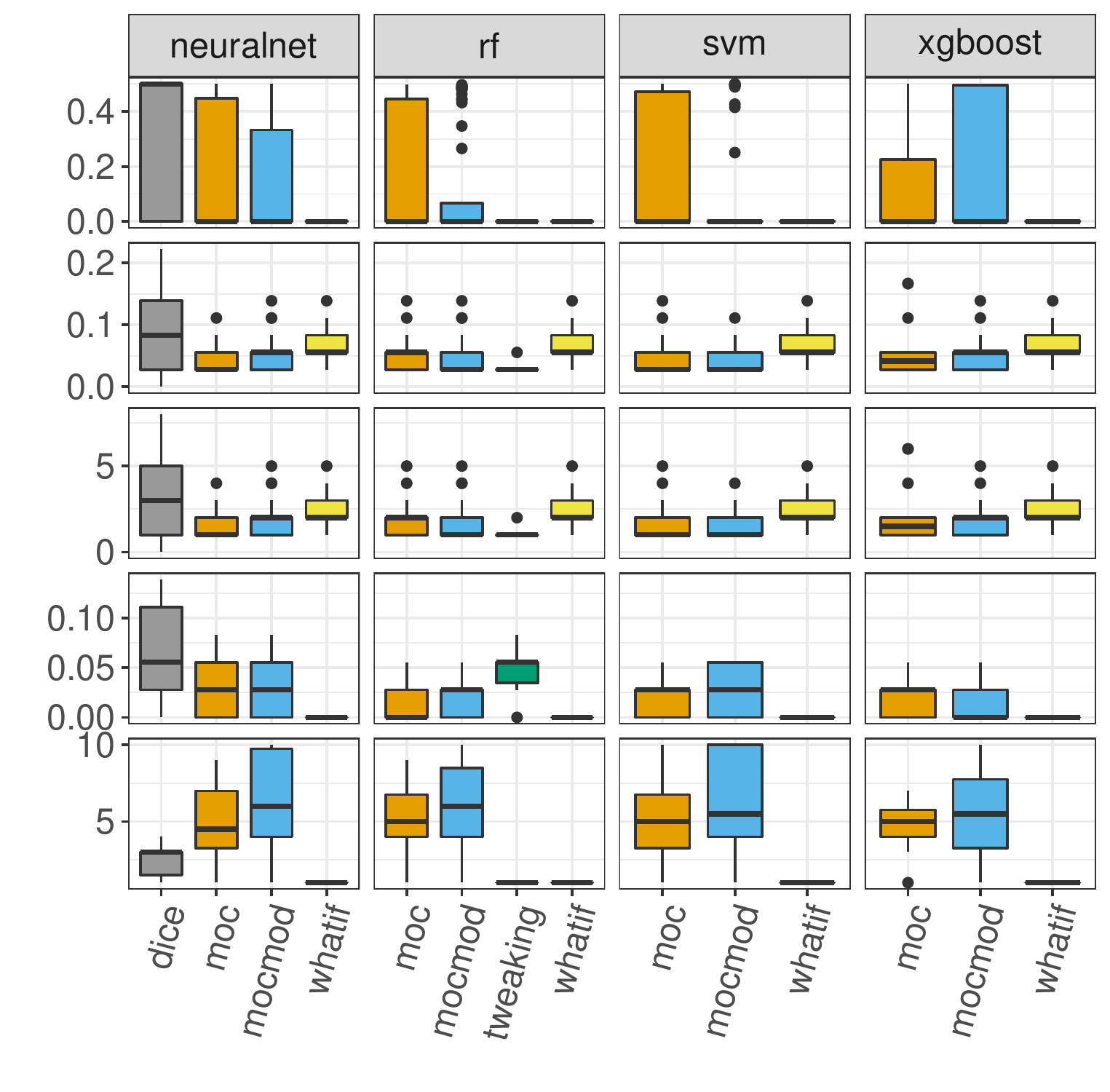}}
	\hspace{-0.38cm}
	\subfigure[cmc]{
		\includegraphics[width=0.49\linewidth]{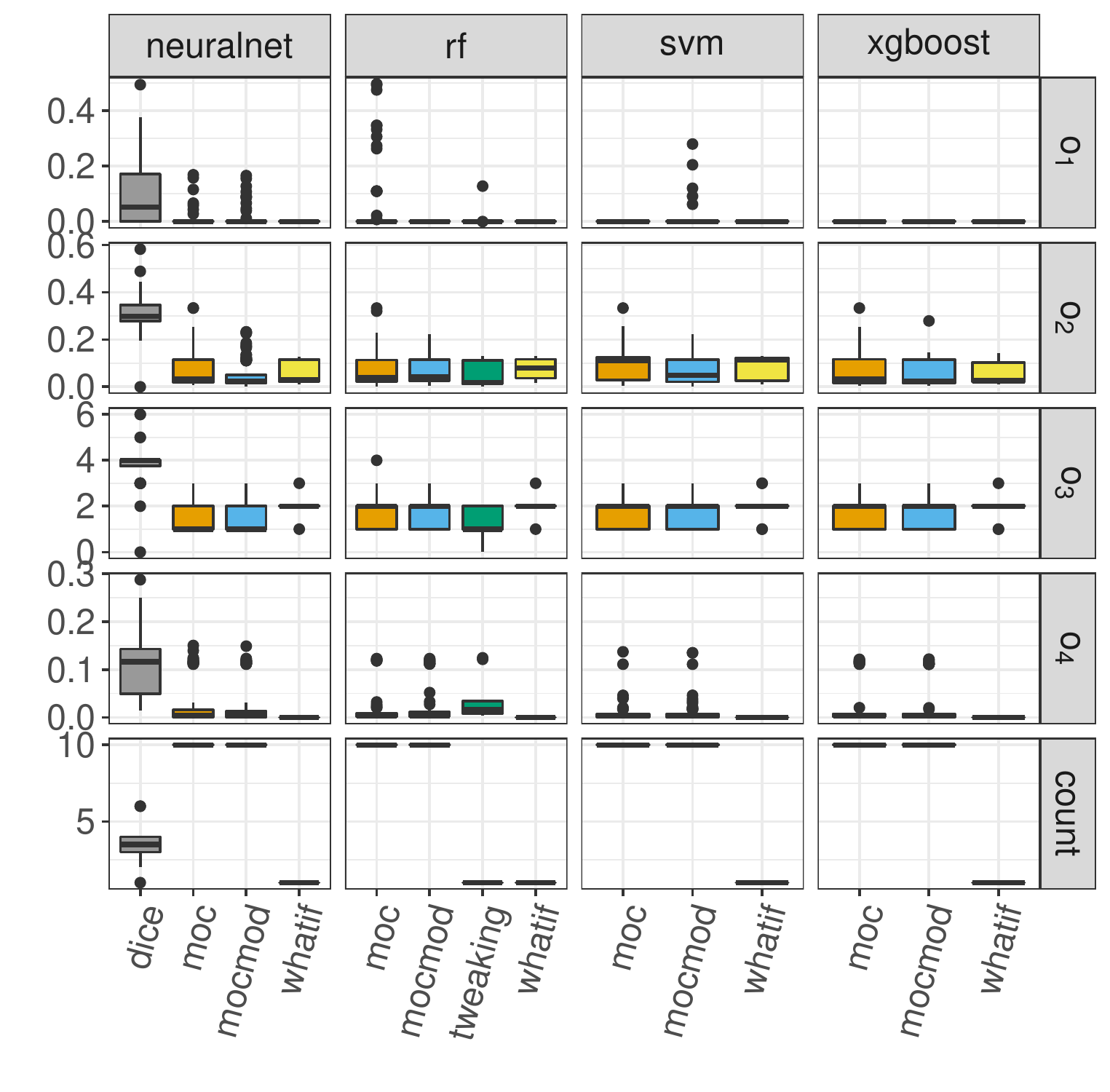}}
	\par\medskip
	\subfigure[plasma\_retinol]{
		\includegraphics[width=0.49\linewidth]{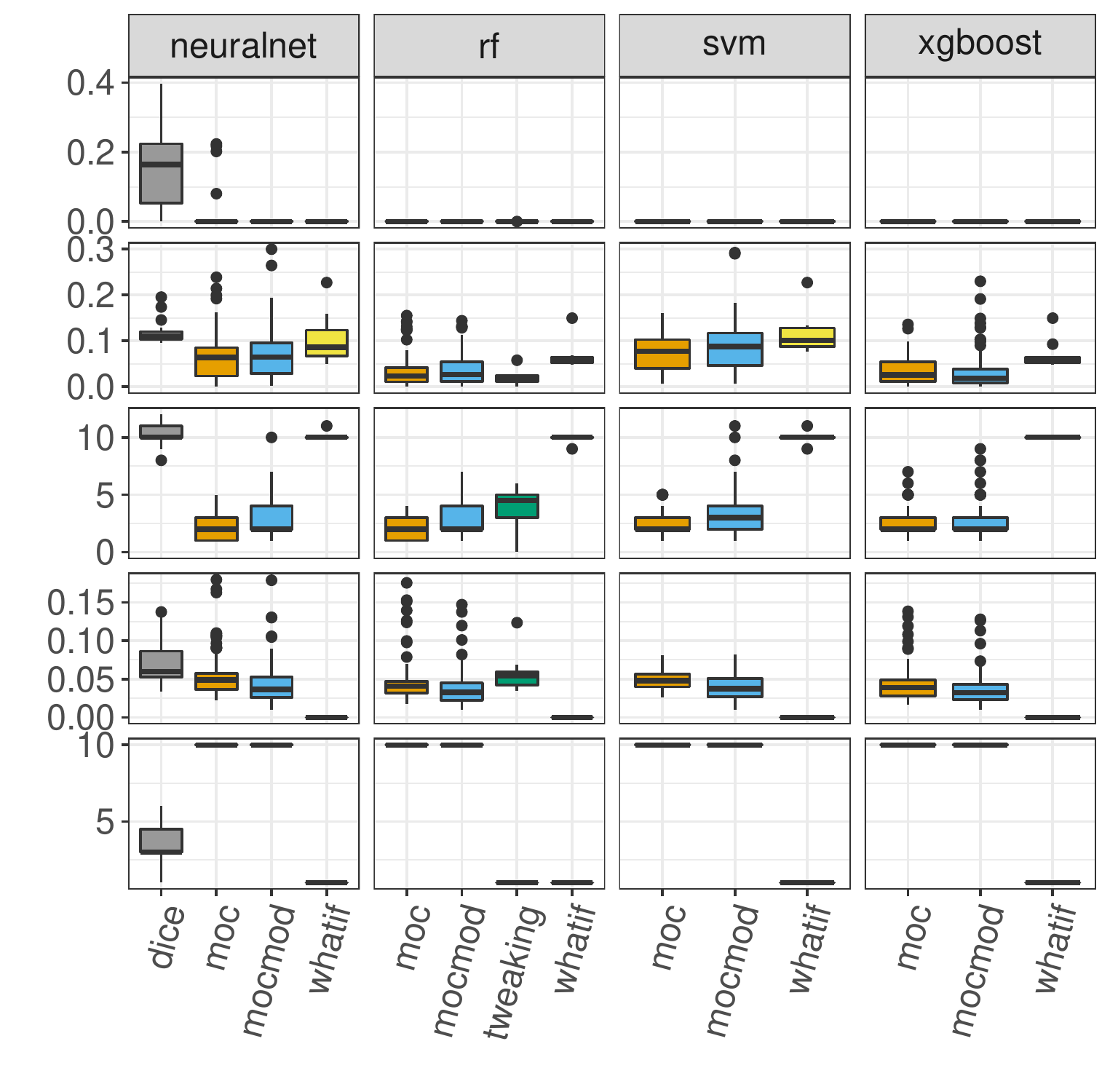}}
	\hspace{-0.38cm}
	\subfigure[tic-tac-toe]{
		\includegraphics[width=0.49\linewidth]{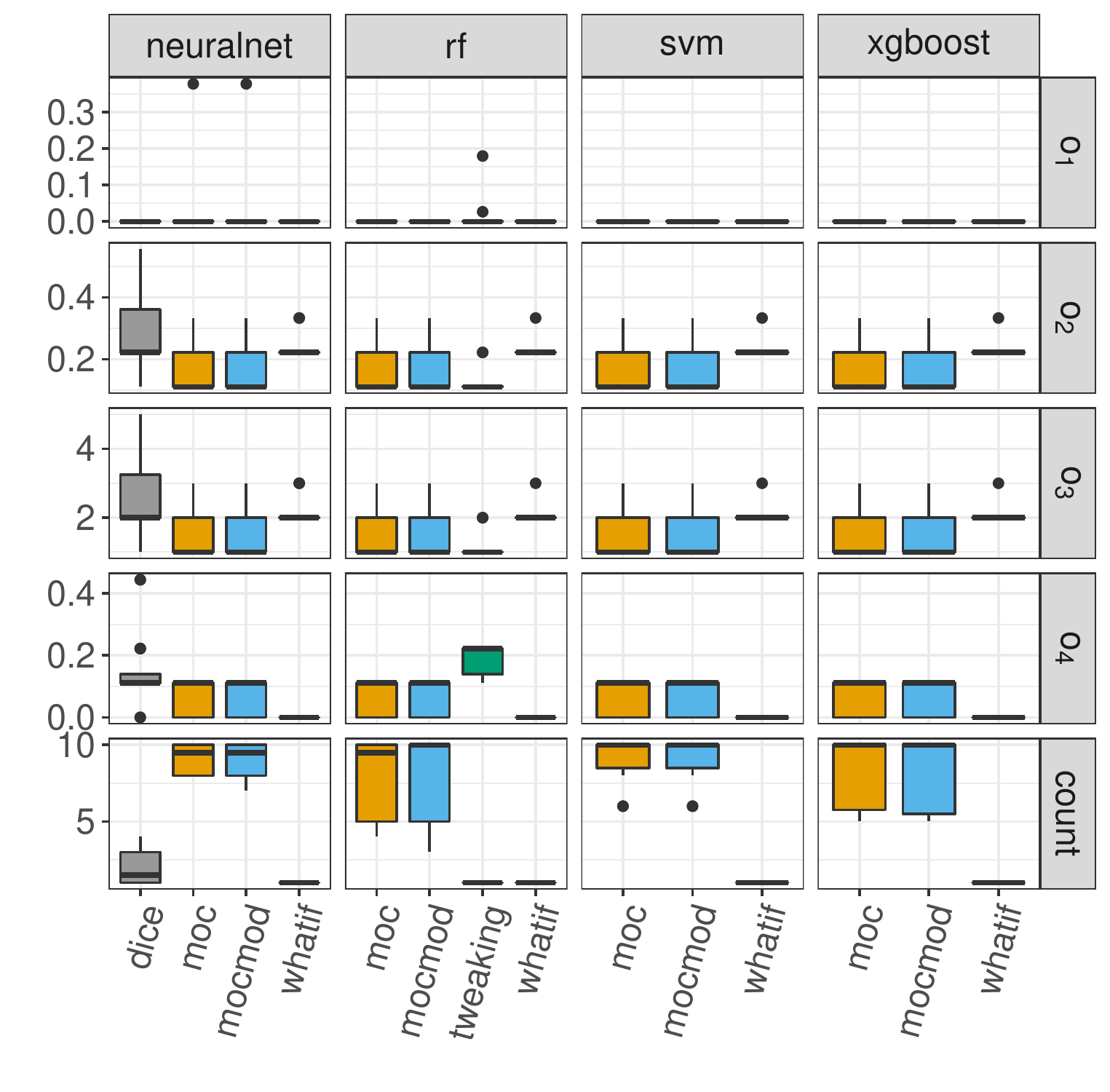}}
	\caption{Boxplots of the objective values and number of nondominated counterfactuals (\textit{count}) per dataset and model for MOC with our proposed strategies for initialization and mutation (\textit{mocmod}), MOC without these modifications, Whatif, DiCE, Recourse and Tweaking. Lower values are better except for \textit{count}.}
	\label{fig:otherboxes2}
\end{figure}

\begin{figure}[ht]
	\centering
	\includegraphics[width=1\textwidth]{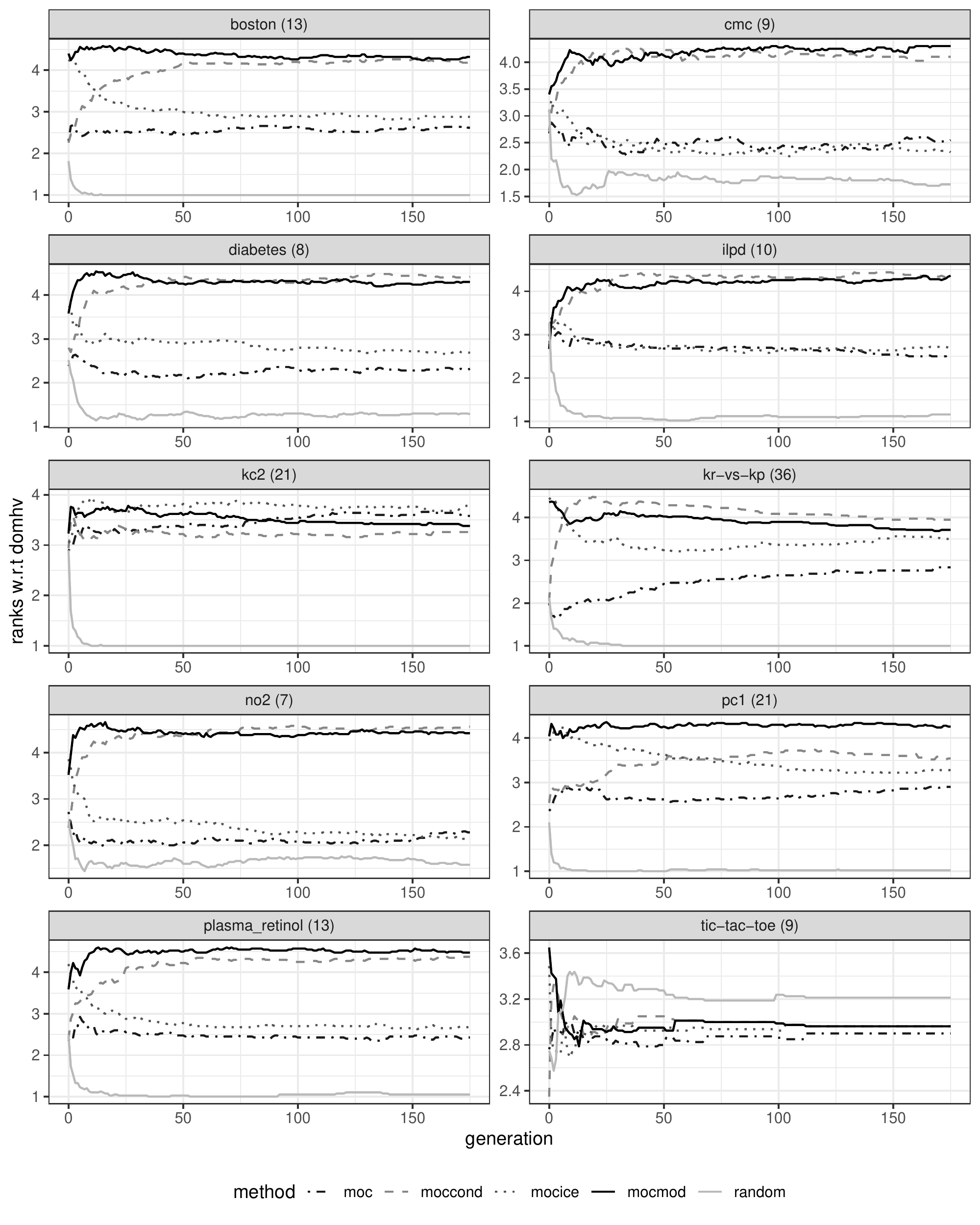}
	\caption{Comparison of the ranks w.r.t. the dominated HV (\textit{domhv}) per generation and per benchmark dataset averaged over all models. The numbers in parentheses indicate the number of features.  For each approach, the population size of each generation was 20. Higher ranks are better. Legend: \textit{moc}: MOC without modifications; \textit{moccond}: MOC with the conditional mutator; \textit{mocice}: MOC with the ICE curve variance initialization; \textit{mocmod}: MOC with both modifications; \textit{random}: random search.}
	\label{ap:hvdataset}
\end{figure} 

\end{document}